\documentclass[lettersize,journal]{IEEEtran}

\usepackage[authoryear,round]{natbib}
\let\cite\citep

\usepackage{caption}
\usepackage{subcaption}
\renewcommand{\thetable}{\arabic{table}}
\makeatletter
\renewcommand\thesection{\arabic{section}}
\renewcommand\thesubsection{\thesection.\arabic{subsection}}
\renewcommand\thesubsubsection{\thesubsection.\arabic{subsubsection}}
\makeatother

\usepackage{amsmath,amsfonts, amssymb}
\usepackage{algorithmic}
\usepackage{algorithm}
\usepackage{array}
\usepackage{textcomp}
\usepackage{stfloats}
\usepackage{url}
\usepackage{verbatim}
\usepackage{graphicx}
\graphicspath{ {./figures/} }
\usepackage[switch]{lineno}
\nolinenumbers
\usepackage{tabularx}
\usepackage{geometry}
\usepackage{float}
\usepackage{tikz}
\usepackage{booktabs}
\usetikzlibrary{3d,decorations.pathreplacing,arrows.meta,positioning,shapes.geometric,fit,calc}
\begin{document}

\title{Post-Disaster Affected Area Segmentation with Vision Transformer (ViT)-based Model using Sentinel-2 and Formosat-5 Imagery}

\author{
    Yi-Shan Chu\IEEEauthorrefmark{1}\IEEEauthorrefmark{2}\IEEEauthorrefmark{3},
    Hsuan-Cheng Wei\IEEEauthorrefmark{2}
    \\
    \IEEEauthorblockA{\IEEEauthorrefmark{1}Department of Mathematical Science, National Chengchi University, Taipei, Taiwan}
    \\
    \IEEEauthorblockA{\IEEEauthorrefmark{2}Division of Satellite Data Application, Taiwan Space Agency, Hsinchu, Taiwan}
    \\
    \IEEEauthorblockA{\IEEEauthorrefmark{3}Institute of Statistical Science, Academia Sinica, Taipei, Taiwan}
    \\
    Email: 111701037@nccu.edu.tw, hsuancheng@tasa.org.tw
}

\maketitle

\begin{abstract}
We propose a vision transformer (ViT)-based deep learning framework to improve disaster-affected area segmentation from satellite images, supporting the Emergent Value Added Product (EVAP) system developed by the Taiwan Space Agency (TASA). The process begins with a small number of manually labeled regions. We then use principal component analysis (PCA) to expand these labels with a confidence interval, creating a weakly supervised training set. Our model, which takes multi-band input from Sentinel-2 and Formosat-5 satellites, is trained to distinguish disaster-affected areas using these expanded labels. We adopt several strategies to increase accuracy when only limited supervision is available. To evaluate performance, our predictions are compared to higher-resolution EVAP results to measure spatial accuracy and consistency. Experiments on the 2022 Poyang Lake drought and the 2023 Rhodes wildfire demonstrate smoother and more reliable delineations. Quantitative evaluation is conducted against manually refined ground truth provided by the Taiwan Space Agency (TASA), with EVAP baseline reported as an operational baseline for comparison.
\end{abstract}


\section{Introduction}
\IEEEPARstart{W}{hen} a disaster occurs, the timely and accurate identification of affected areas is crucial for guiding emergency response and mitigating further losses. To support this need, the Taiwan Space Agency (TASA) developed the Emergent Value-Added Product (EVAP) system—a semi-automated geospatial workflow designed to assist in rapid disaster mapping once an event has been reported and verified~\cite{hung2024evap}. EVAP utilizes a combination of spectral indices such as the Normalized Difference Vegetation Index (NDVI) and the Normalized Difference Water Index (NDWI), and change vector analysis to detect changes between pre- and post-disaster remote sensing imagery. A supervised Gaussian statistical method is then employed, requiring analysts to manually label a small number of disaster-affected polygons (typically fewer than ten), which are used to define confidence intervals and classify affected regions across the entire image. 

Although systems such as NASA's Disaster Mapping Dashboard and UNOSAT provide post-disaster assessments, they often rely on human interpretation or high-resolution commercial satellite imagery. In contrast, EVAP offers a semi-automated and resource-efficient alternative that takes advantage of freely available satellite data. However, it currently lacks the capacity for deep learning-based generalization, which limits its scalability and adaptability across diverse disaster scenarios.

While EVAP has demonstrated operational efficiency across diverse disaster scenarios, its reliance on user-defined training samples and Gaussian distribution assumptions can limit its adaptability and accuracy, particularly in complex or heterogeneous environments. Moreover, the quality of its output is closely tied to the resolution and spectral characteristics of the input imagery, which can vary significantly across satellite platforms. In addition, EVAP employs a pixel-wise statistical classification procedure which, although effective at small scales, becomes computationally expensive when processing large-scale satellite imagery. As the spatial coverage and resolution increases, the pixel-wise computation lead to long processing times and poses challenges for timely disaster response in operational settings. 

At the same time, Vision Transformers (ViTs)~\cite{dosovitskiy2020vit} have become increasingly popular in remote sensing tasks due to their ability to model long-range spatial relationships and capture global context more effectively than traditional convolutional neural networks (CNNs). ViT-based architectures have shown strong performance in semantic segmentation tasks involving high-resolution aerial and satellite imagery, frequently outperforming conventional CNNs.

Moreover, ViT-based methods have also been widely applied for change detection tasks. Prominent models such as ChangeFormer~\cite{bandara2022changeformer}, ChangeViT~\cite{zheng2023changevit}, and Siamese ViT frameworks~\cite{li2022siamvit} have achieved state-of-the-art results on public datasets like LEVIR-CD~\cite{chen2020levircd} and xBD~\cite{gupta2019xbd}. However, these approaches typically assume access to very high-resolution (VHR), mono-source imagery and rely on fully supervised training with pixel-level ground truth annotations. Such conditions are rarely available in time-critical or resource-constrained disaster response settings.

Training deep models under weak supervision—especially when labels are derived from heuristic or low-resolution outputs—remains a major challenge in remote sensing. Prior ViT-based methods often overlook these constraints, limiting their applicability to real-world operational systems. Investigating whether such supervision can still produce reliable and generalizable segmentation is both scientifically non-trivial and practically valuable.

In this work, we aim to bridge this gap by adapting ViT-based segmentation to enhance EVAP under real-world constraints. Specifically, we target large-scale disaster-affected region segmentation using medium-resolution, multi-source satellite imagery from Sentinel-2 and Formosat-5, where supervision is provided only through low-resolution EVAP outputs. We explore how transformer-based deep learning models can improve the spatial consistency and generalization ability of EVAP, offering a scalable upgrade to current operational pipelines for disaster impact mapping.

The main contributions of this work are:
\begin{enumerate}
    \item We adapt Vision Transformer-based segmentation models to the context of medium-resolution, multi-source disaster imagery with weak supervision.

    \item We introduce a principal component analysis (PCA)- and confidence interval (CI)-based label expansion strategy, which enhances weak supervisory signals derived from low-resolution EVAP outputs.

    \item We validate our approach on multiple disaster case studies using Sentinel-2 and Formosat-5 imagery, demonstrating improvements in spatial coherence and inference efficiency compared to the original EVAP method.
\end{enumerate}

\section{Related Works}
\subsection{Disaster-Affected Area Segmentation}
Accurate semantic segmentation of disaster-affected regions in remote sensing imagery is crucial for rapid damage assessment. Traditional techniques often relied on spectral indices or simple thresholding to delineate affected areas (e.g., using NDWI for floods or Normalized Burn Ratio (NBR) for burn scars), but recent deep learning models have substantially improved segmentation accuracy. For instance, \citet{fakhri2025Flood} applied an attention-based U-Net to Sentinel-1 Synthetic Aperture Radar (SAR) images for flood mapping, achieving high precision and recall ($\sim$0.90) in delineating flooded regions. Their attention-based model could extract water inundation areas from post-flood SAR scenes. 

In landslide segmentation, \citet{li2023landslide} propose an improved U-Net architecture with dilated convolutions and an efficient multiscale attention (EMA) mechanism, which enhances the extraction of features for landslide scars. By redesigning the encoder and introducing a novel skip-connection module, their model outperformed the vanilla U-Net by $\sim$2--3\% in mIoU and F1-score. Wildfire damage mapping has similarly benefited from tailored CNN architectures: \citet{khankeshizadeh2024} develop a dual-path attention residual U-Net that fuses multispectral optical and SAR imagery to segment burned areas. Experiments in multiple wildfire cases showed that the approach achieved IoU up to 89.3\%, outperforming conventional U-Net baselines.

\subsection{Post-Disaster Change Detection}
Beyond single-image analysis, many works perform change detection using pre- and post-disaster image pairs to identify affected areas. Deep learning has replaced earlier pixel-wise change detection techniques (e.g., change vector analysis) with learned representations that better distinguish true damage from irrelevant changes (seasonal differences, shadows, etc.). Modern change detection networks typically adopt a Siamese or encoder–decoder architecture to process inputs (e.g., \citealp{chen2017rethinking, chen2020spatial, xview2challenge}).

\subsection{Vision Transformers in Remote Sensing}
The application of Vision Transformers in remote sensing has driven state-of-the-art results in both segmentation and change detection tasks. Examples include UNetFormer for semantic segmentation \cite{Wang2022UNetFormer}, ChangeFormer for change detection \cite{bandara2022changeformer}, and FTN \cite{Yan2022FTN}, which report state-of-the-art accuracy by capturing global context and reducing false alarms.

\subsection{Weak Supervision in Remote Sensing Segmentation}
Weakly and semi-supervised techniques reduce dense labeling requirements. Representative approaches include CAM-based pseudo masks \cite{Cao2023Weak, Lu2024Weak}, SAM-assisted proposals \cite{Chen2025Siamese, Kirillov2023SAM}, and pseudo-label self-training \cite{Wang2021ALS}. Our approach differs by using PCA with confidence intervals to expand seed labels, avoiding model-generated pseudo labels.

\subsection{Emergent Value-Added Product (EVAP)}
Many space agencies routinely generate value-added products (VAPs) to support rapid disaster response. The EVAP workflow commonly uses statistical thresholds on indices and change measures, including change vector analysis, to highlight affected zones. Recent work introduces Gaussian-mixture-based confidence bounds to improve robustness \cite{Chung2023EVAP}.

\begin{figure*}[ht]
    \centering
    \begin{tikzpicture}[scale=0.82, every node/.style={scale=0.97}]
        \foreach \i/\c in {0/blue!20, 0.13/blue!30, 0.26/blue!40, 0.39/blue!60} {
            \draw[thick, fill=\c, opacity=0.90-\i*1.2] (\i,\i) rectangle (2+\i,2+\i);
        }
        \draw[decorate,decoration={brace,amplitude=6pt},xshift=-7pt] (0,0) -- (0,2);
        \node at (-0.75,1) {\footnotesize $H$};
        \draw[decorate,decoration={brace,mirror,amplitude=6pt},yshift=-5pt] (0,0) -- (2,0);
        \node at (1,-0.7) {\footnotesize $W$};
        \node at (1.12,2.7) {\footnotesize \textbf{4-band}};
        \node at (1,-1.2) {\footnotesize \textbf{Pre-disaster}};
        \node at (1,-1.8) {\footnotesize $I_{\mathrm{pre}}$};
        \node[font=\large] at (2.52,1) {+};

        \foreach \i/\c in {0/red!20, 0.13/red!30, 0.26/red!40, 0.39/red!60} {
            \draw[thick, fill=\c, opacity=0.90-\i*1.2] (2.7+\i,\i) rectangle (4.7+\i,2+\i);
        }
        \node at (3.82,2.7) {\footnotesize \textbf{4-band}};
        \node at (3.7,-1.2) {\footnotesize \textbf{Post-disaster}};
        \node at (3.7,-1.8) {\footnotesize $I_{\mathrm{post}}$};

        \node[font=\large] at (5.35,1) {=};

        \foreach \i/\c in {0/purple!20, 0.10/purple!30, 0.20/purple!40, 0.30/purple!60, 0.40/purple!40, 0.50/purple!30, 0.60/purple!20, 0.70/purple!10} {
            \draw[thick, fill=\c, opacity=0.7] (5.6+\i,\i) rectangle (7.6+\i,2+\i);
        }

        \node at (7.22,3) {\footnotesize \textbf{8-band}};
        \node at (7,-1.2) {\footnotesize \textbf{Stack Input}};
        \node at (7,-1.8) {\footnotesize $X$};

        \draw[thick,->] (8.5,1) -- (9.1,1);

        \node[draw,thick,fill=white!95!gray!30,rounded corners=2pt, minimum width=1.3cm, minimum height=0.6cm] at (10,1) {Model};

        \draw[thick,->] (10.9,1) -- (11.5,1);

        \draw[thick,fill=gray!15] (11.7,0) rectangle (13.7,2);
        \node[align=center] at (12.7,-0.6) {\footnotesize \textbf{Output Mask}};
        \node[align=center] at (12.7,-1.2) {{\footnotesize $Y \in \{0,1\}^{H\times W}$}};
    \end{tikzpicture}
    \caption{Schematic diagram illustrating the construction of the input tensor $X$ by concatenating pre-disaster ($I_{\mathrm{pre}}$) and post-disaster ($I_{\mathrm{post}}$) multi-band images along the channel dimension.}
    \label{fig:input-construction}
\end{figure*}

\section{Proposed Method}

\subsection{Problem Setup}

Our objective is to segment disaster-affected regions using multi-temporal remote sensing imagery acquired from Sentinel-2~\cite{sentinel2} and Formosat-5~\cite{formosat5}. For each target area, we acquire pre-disaster and post-disaster images, each with four spectral bands (R, G, B, NIR). To facilitate joint analysis, both images are co-registered and resampled to a common spatial resolution. As illustrated in Fig. \ref{fig:input-construction}, the resulting input can be represented as an 8-channel image array:
\[
X = [I_{\mathrm{pre}}; I_{\mathrm{post}}] \in \mathbb{R}^{H \times W \times 8}
\]
where $I_{\mathrm{pre}}, I_{\mathrm{post}} \in \mathbb{R}^{H \times W \times 4}$ are the pre- and post-disaster images. The segmentation task is to predict a binary mask $Y \in \{0,1\}^{H \times W}$, indicating the disaster-affected areas.

Multi-satellite integration introduces challenges such as differing spectral responses and radiometric characteristics. Furthermore, the medium spatial resolution of Sentinel-2 may fail to resolve fine-scale features, making robust modeling strategies essential for accurate segmentation.

\subsection{Labeling Strategy}
In scenarios where disaster causes substantial changes to the landscape, we hypothesize that pixels undergoing dramatic change will form a coherent cluster in the projected feature space. Therefore, our label expansion strategy utilizes this assumption, allowing us to incorporate pixels with high similarity, i.e., those lying within the confidence interval under Gaussian distribution, as additional positive samples. This assumption is supported by the observation that disaster-affected pixels often exhibit consistent changes in spectral and principal component space.

Given the limited availability of high-quality, manually annotated masks, we employ a semi-automatic labeling strategy to generate training data efficiently. Initially, a small region $\mathcal{A} \subset X$ are manually annotated as affected regions. The 8-dimensional spectral vectors at these locations are used as the positive class for further expansion.

To enhance label coverage and reduce dimensionality, we perform principal component analysis (PCA) on the concatenated spectral features and project all pixels into a reduced $k$-dimensional space:
\begin{equation}
P = \mathrm{PCA}_k(X)
\end{equation}
Assuming that the positive samples form an approximate Gaussian cluster in PCA space, we compute the mean $\mu$ and covariance $\Sigma$ from the seed set, and define a confidence region using the Mahalanobis distance:
\begin{equation}
d_M(p) = \sqrt{(p - \mu)^\top \Sigma^{-1} (p - \mu)}
\end{equation}
For a given confidence level $\alpha$ (e.g., $\alpha=0.99$), the corresponding Mahalanobis distance threshold $\tau$ is determined such that
\begin{equation}
\tau^2 = \chi^2_{k,\alpha}
\end{equation}
where $\chi^2_{k,\alpha}$ is the upper $\alpha$-quantile of the chi-squared distribution with $k$ degrees of freedom.  
All pixels whose projected feature vectors satisfy $d_M(p) < \tau$ are assigned as additional positive labels:
\begin{equation}
    \mathcal{L} = \mathcal{A} \cup \{ (i,j) \in \Omega \setminus \mathcal{A} \mid d_M(P_{i,j}) < \tau \}
\end{equation}
where $\mathcal{A}$ is the set of initial seed pixels, $\Omega$ is the set of all pixel coordinates, and $\mathcal{L}$ is the expanded labeled set. This enables weak supervision at scale with minimal manual intervention.

\subsection{Model Architecture}
\begin{figure*}[ht]
    \centering
    \includegraphics[width=0.99\linewidth]{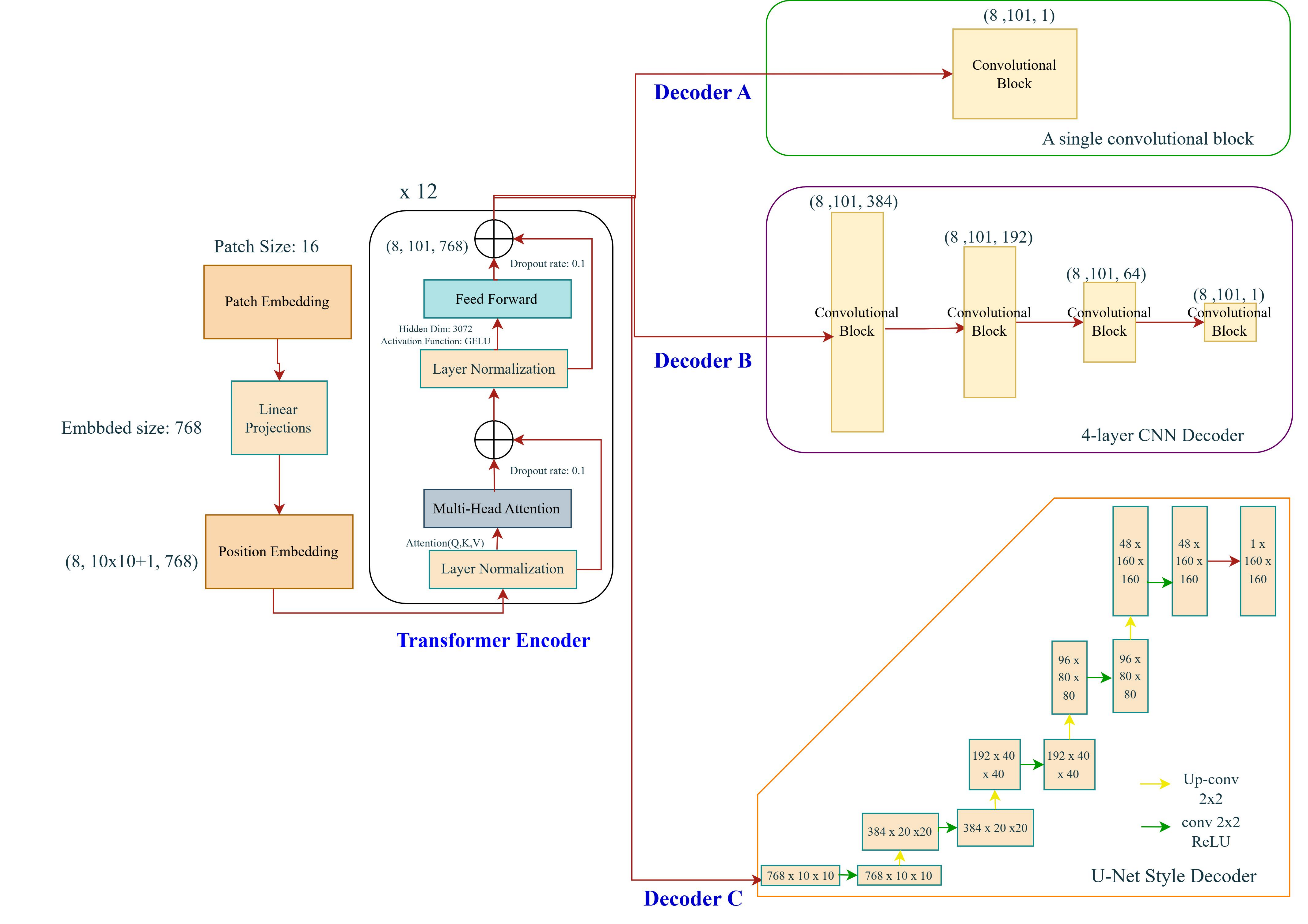}
    \caption{Comparison of model architectures used in this work. \textbf{A}: Vision Transformer (ViT) encoder with single-block decoder. \textbf{B}: ViT encoder with 4-layer CNN decoder. \textbf{C}: ViT encoder with U-Net style decoder.}
    \label{fig:model-arch}
\end{figure*}

Our deep learning framework adopts a modular encoder–decoder structure for disaster-affected area segmentation, as illustrated in Fig.~\ref{fig:model-arch}. Specifically, all models share a common encoder design based on the Vision Transformer (ViT), while differing in the design of the decoder. This design allows us to investigate the impact of various decoder architectures on segmentation performance under weak supervision.

\textbf{(a) ViT Encoder:}
The encoder follows the standard Vision Transformer paradigm, partitioning the input image into non-overlapping patches, linearly embedding each patch, and processing the resulting sequence with transformer blocks. The encoder extracts high-level, non-local features from the multi-band input, enabling the model to capture complex disaster-induced changes.

\textbf{(b) Decoders:}
We evaluate three decoder architectures:
\begin{itemize}
    \item \textbf{Decoder A: Single-block convolutional decoder.} This minimalistic decoder consists of a single convolutional block applied to the ViT-encoded features, projecting them directly to the output binary mask. It serves as a lightweight baseline for comparison.
    \item \textbf{Decoder B: Multi-layer CNN decoder.} This variant employs a four-layer convolutional neural network (CNN) decoder, progressively upsampling and refining the feature maps to recover spatial resolution and detail.
    \item \textbf{Decoder C: U-Net style decoder.} Inspired by the U-Net architecture, this decoder incorporates symmetric upsampling and skip connections, which help preserve fine-grained spatial information and enable robust segmentation of small or fragmented affected regions.
\end{itemize}

\subsection{Loss Functions}
To enable robust learning under weak supervision, we employ three different loss functions for training our segmentation models: (1) Binary Cross Entropy (BCE), (2) BCE-Dice Loss, and (3) BCE-IoU Loss. The third loss adopts a two-stage training approach, where the model is first trained to convergence with BCE loss, and then further fine-tuned using the IoU loss.

\textbf{(1) Binary Cross Entropy (BCE):}
\begin{equation}
    \mathrm{BCE}(\mathbf{x}, \mathbf{y}) = -\frac{1}{N} \sum_{i=1}^N \left[ y_i \log x_i + (1 - y_i) \log (1 - x_i) \right]
\end{equation}

\textbf{(2) BCE-Dice Loss:}
\begin{align}
    \mathrm{BCE\text{-}Dice}(\mathbf{x}, \mathbf{y}) &= \mathrm{BCE}(\mathbf{x}, \mathbf{y}) + \mathrm{Dice}(\mathbf{x}, \mathbf{y}) \\
    \mathrm{Dice}(\mathbf{x}, \mathbf{y}) &= 1 - \frac{2 \sum_{i=1}^N x_i y_i}{\sum_{i=1}^N x_i + \sum_{i=1}^N y_i}
\end{align}

\textbf{(3) BCE-IoU Loss (Two-Stage):}
\begin{align}
    \mathrm{IoU}(\mathbf{x}, \mathbf{y}) &= 1 - \frac{\sum_{i=1}^N x_i y_i}{\sum_{i=1}^N \left(x_i + y_i - x_i y_i\right)}
\end{align}

For BCE-IoU (two-stage Approach), we first optimize the model using the BCE loss until convergence (i.e., until the validation loss plateaus for epochs), after which the training continues using the IoU loss for further refinement.

\noindent
Here, $N$ denotes the total number of pixels, $x_i$ is the predicted value for the $i$-th pixel, and $y_i$ is the corresponding ground-truth label (0 or 1). This multi-loss framework ensures that the models not only achieve accurate pixel-wise classification but also better capture the spatial structure of disaster-affected areas.

\section{Dataset}
To evaluate our approach, we consider two real-world disaster scenarios using multi-sensor remote sensing data. We utilize images from two complementary satellite platforms: Sentinel-2~\cite{sentinel2} and Formosat-5~\cite{formosat5}.

\subsection{Data Sources}
\subsubsection{(a) Sentinel-2}
Sentinel-2 (S2), operated by the European Space Agency (ESA), is a constellation of twin satellites launched in 2015 and 2017. Each Sentinel-2 satellite carries a MultiSpectral Instrument (MSI) capable of capturing 13 spectral bands ranging from visible to shortwave infrared, at spatial resolutions of 10\,m, 20\,m, and 60\,m depending on the band. The satellite provides a global revisit time of 5 days, making it well-suited for monitoring rapid environmental changes and disasters.

\subsubsection{(b) Formosat-5}
Formosat-5 (FS5) is Taiwan's first independently developed remote sensing satellite, launched in 2017 by the National Space Organization (NSPO), which has since been reorganized as the Taiwan Space Agency (TASA). The satellite is equipped with an optical payload that acquires images in four bands (red, green, blue, and near-infrared) with a ground sampling distance of 2\,m (panchromatic) and 4\,m (multispectral). Formosat-5 is designed for applications in disaster monitoring, environmental assessment, and land use mapping.

\begin{table*}[htbp]
\label{tab:sat-comparison}
\centering
\begin{tabular}{lcc}
\hline
\textbf{Specification} & \textbf{Sentinel-2} & \textbf{Formosat-5} \\
\hline
Operator         & ESA                    & TASA (formerly NSPO) \\
Launch Year      & 2015 (S2A), 2017 (S2B) & 2017                 \\
Spectral Bands   & 13                     & 4                    \\
Spatial Resolution & 10\,m / 20\,m / 60\,m & 2\,m (PAN), 4\,m (MS) \\
Swath Width      & 290 km                 & 24 km                \\
Revisit Time     & 5 days                 & 2 days (Taiwan), $\sim$1 week (global) \\
Main Applications & Land monitoring, disaster, agriculture & Disaster, environment, land use \\
Data Access      & Public                 & Public               \\
\hline
\end{tabular}
\caption{Comparison of Sentinel-2 and Formosat-5 satellites (PAN: panchromatic; MS: multispectral).}
\end{table*}

\subsection{Case Studies}  
Two disaster scenarios considered in this study:

\begin{itemize}
    \item \textbf{2023 Rhodes Wildfire.}
    Pre- and post-disaster images are collected over Rhodes, Greece, which suffered severe wildfires in July 2023. The pre-disaster image is a Sentinel-2 acquisition from July 1, 2023, while the post-disaster image is a Formosat-5 acquisition from August 1, 2023.
    \item \textbf{2022 Poyang Lake Drought.}
    To study large-scale hydrological change, we select Poyang Lake, China, which experienced significant drought in 2022. The pre-disaster image is a Sentinel-2 acquisition from May 16, 2022, and the post-disaster image is a Formosat-5 acquisition from September 2, 2022.
\end{itemize}

For both cases, the Red, Green, Blue, Near Infrared bands are extracted, resampled, and co-registered to a common spatial grid. The combination of Sentinel-2’s medium-resolution multispectral data with Formosat-5’s high-resolution imagery enables robust assessment of our proposed segmentation and label expansion methods under diverse disaster scenarios.

\section{Experiment Results}

\subsection{Experimental Workflow}
\begin{figure*}[ht]
    \centering
    \includegraphics[width=0.97\linewidth]{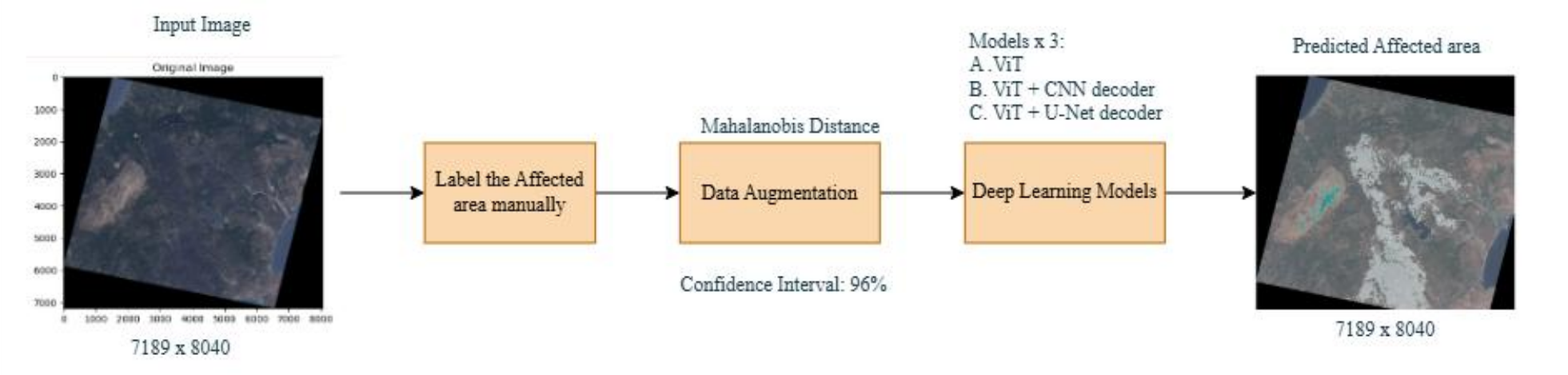}
    \caption{Proposed system pipeline for disaster-affected area segmentation. The workflow consists of initial manual annotation, label expansion using Mahalanobis distance in the PCA feature space, followed by training of deep learning segmentation models.}
    \label{fig:sys-architecture}
\end{figure*}

\begin{figure*}[ht]
    \centering
    \includegraphics[width=0.97\linewidth]{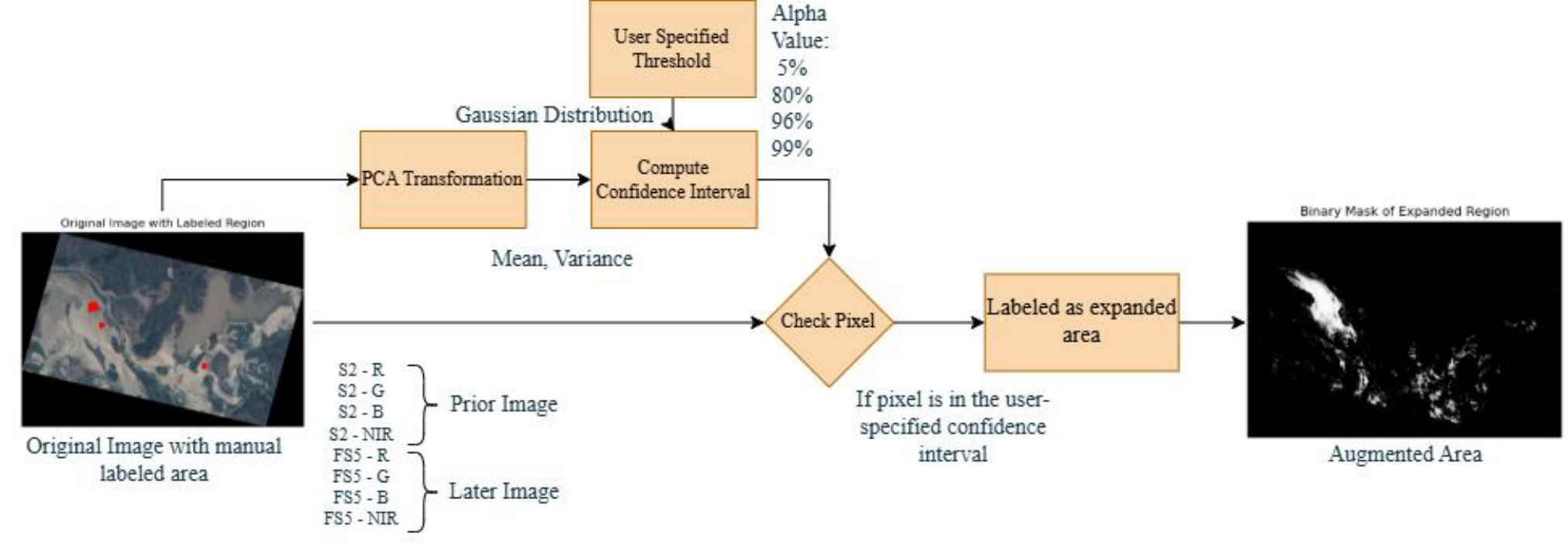}
    \caption{Illustration of the label expansion pipeline. Manually labeled seed regions are projected into a reduced feature space via PCA. Pixels falling within a high-confidence region (as determined by Mahalanobis distance and user-specified confidence interval) are automatically assigned as expanded positive samples, producing an augmented label mask for weakly supervised learning.}
    \label{fig:label-expansion}
\end{figure*}

The overall experimental workflow is illustrated in Fig.~\ref{fig:sys-architecture}. The process begins with the collection of pre- and post-disaster satellite imagery, followed by manual annotation of affected regions. To address label scarcity, we employ a semi-automatic label expansion technique based on Mahalanobis distance in the PCA feature space, as detailed in Section 3 and shown in Fig.~\ref{fig:label-expansion}. The augmented label masks are then used to train multiple segmentation models.


\subsection{Patch Extraction and Data Preparation}
Because high-resolution remote sensing images are too large to be processed by deep learning models in a single pass, we extract fixed-size patches for both training and inference. Specifically, each image is divided into non-overlapping patches of size $H_p \times W_p$ (e.g., $256 \times 256$ pixels). This patch-based approach allows for efficient utilization of GPU memory and enables local context modeling. For evaluation, the predicted patch-wise outputs are reassembled into full-scene masks. This pre-processing step is critical for both model convergence and computational feasibility.

\subsection{Model Architectures and Training}
The full model architectures and training strategy are summarized in Fig.~\ref{fig:f-model-arch}. All variants employ a Vision Transformer (ViT) encoder, with one of three decoder designs: (A) a single-block convolutional decoder, (B) a four-layer CNN decoder, or (C) a U-Net-style decoder. Each model is trained with multiple loss functions (see Section~3.4), and training is performed on four NVIDIA Tesla V100-32G GPUs to accommodate the large tiff images and model size.

\begin{figure*}[ht]
\centering
\begin{tikzpicture}[
    font=\small,
    node distance=3mm and 6mm,
    >=LaTeX,
    block/.style={draw, rounded corners, align=center, fill=white, minimum width=14mm, minimum height=7mm, inner sep=2pt},
    io/.style={draw, align=center, fill=white, minimum width=14mm, minimum height=7mm, inner sep=2pt},
    loss/.style={draw, dashed, rounded corners, align=left, fill=white, minimum width=24mm, inner sep=3pt},
    gt/.style={draw, thick, align=center, fill=white, minimum width=34mm, minimum height=7mm, inner sep=2pt},
    note/.style={draw, densely dotted, rounded corners, align=left, fill=white, inner sep=3pt},
    merge/.style={draw, circle, inner sep=0pt, minimum size=2mm},
    shorten >=1pt, shorten <=1pt
]

\node[block] (stack) {Stacked Input\\ $X=[I_{\text{pre}};I_{\text{post}}]$};

\node[block, right=3mm of stack] (vit) {\textbf{ViT encoder}};

\node[block, above right=4mm and 2mm of vit] (decA) {\textbf{Decoder A}\\ single conv};
\node[block, right=3mm of vit] (decB) {\textbf{Decoder B}\\ 4-layer CNN};
\node[block, below right=4mm and 2mm of vit] (decC) {\textbf{Decoder C}\\ U-Net style};

\node[block, right=3mm of decA, minimum width=14mm] (predA) {Prediction};
\node[block, right=3mm of decB, minimum width=14mm] (predB) {Prediction};
\node[block, right=3mm of decC, minimum width=14mm] (predC) {Prediction};

\node[loss, right=5mm of predB, yshift=1mm] (lossbox) {\textbf{Training losses}\\
(1) Binary Cross Entropy (BCE)\\
(2) BCE--Dice\\
(3) Two-stage BCE$\rightarrow$IoU};

\node[gt, below=6mm of lossbox] (gtbox) {\textbf{Ground Truth}\\(expert annotation)};

\node[block, right=3mm of lossbox, yshift=-1mm, align=left, minimum width=15mm] (eval) {\textbf{Evaluation}};

\draw[->] (stack) -- (vit);

\draw[->] (vit) |- (decA);
\draw[->] (vit) -- (decB);
\draw[->] (vit) |- (decC);

\draw[->] (decA) -- (predA);
\draw[->] (decB) -- (predB);
\draw[->] (decC) -- (predC);

\draw[->] (predA) -| (lossbox.north);
\draw[->] (predB) -- (lossbox);
\draw[->] (predC) -| (lossbox.south);

\draw[->, thick] (gtbox.north) -- (lossbox.south);
\draw[->] (lossbox) -- (eval);

\end{tikzpicture}

\caption{Flowchart of the experimental pipeline. Pre/post RGBN (S2/FS5) are stacked and fed into a ViT encoder with three decoder variants: \textbf{A} (single conv), \textbf{B} (4-layer CNN), \textbf{C} (U-Net style). Models are trained under three losses: (1) BCE, (2) BCE--Dice, (3) two-stage BCE$\rightarrow$IoU. \textbf{All metrics are computed against manually refined ground truth(GT).}}
\label{fig:f-model-arch}
\end{figure*}

\begin{figure*}[htbp]
  \centering
  \includegraphics[width=\textwidth]{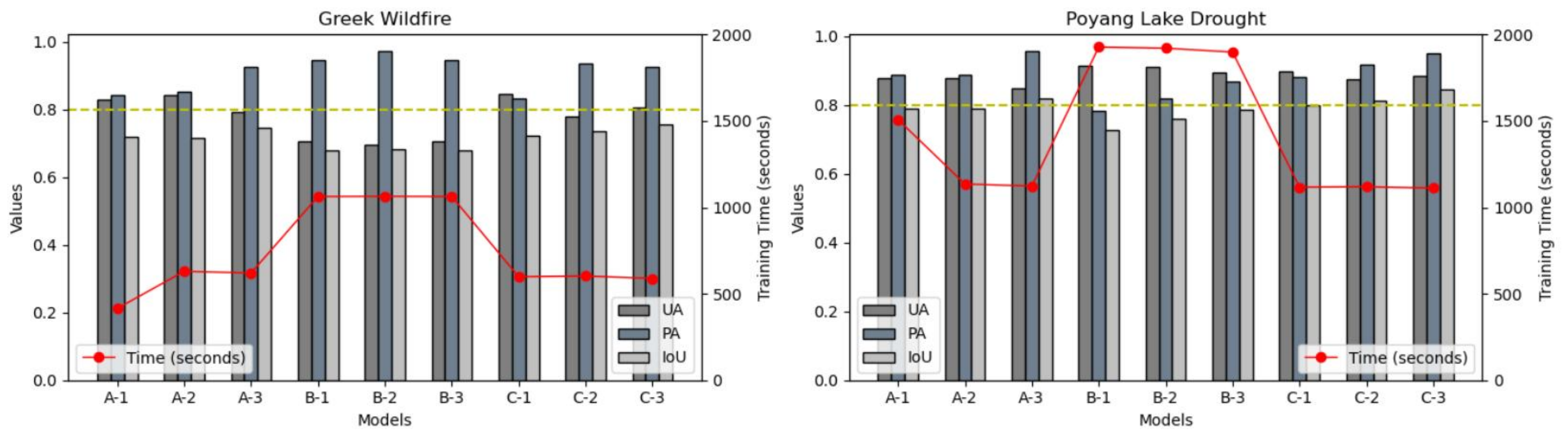}
  \caption{Quantitative evaluation on the Greek Wildfire and Poyang Lake Drought datasets. The bar plots show the UA, PA, and IoU metrics for various model configurations, while the red line indicates training time in seconds. A/B/C denote decoder variants with increasing complexity: (A) with a single convolution block, (B) with a 4-layer CNN, and (C) with a U-Net decoder. The numeric suffix (1/2/3) refers to different loss settings: (1) BCE loss, (2) BCE-Dice loss, and (3) 2-stage loss. Models are trained on 4× Tesla V100 (32GB) GPUs.}
  \label{fig:exp_results}
\end{figure*}

\subsection{Quantitative Results and Metrics}

\noindent We evaluate model performance using three widely adopted segmentation metrics: User Accuracy (UA), Producer Accuracy (PA), and Intersection over Union (IoU). Given the set of predicted positive pixels $P$ and ground truth positive pixels $G$, these metrics are defined as follows:
\begin{equation}
  \mathrm{UA} = \frac{|P \cap G|}{|P|}
\end{equation}
\begin{equation}
  \mathrm{PA} = \frac{|P \cap G|}{|G|}
\end{equation}
\begin{equation}
  \mathrm{IoU} = \frac{|P \cap G|}{|P \cup G|}
\end{equation}

\begin{table}[t]
\centering
\begin{tabular}{lccc}
\toprule
\multicolumn{4}{c}{\textbf{Wildfire case}}\\
\midrule
Method & PA & UA & IoU \\
\midrule
\textbf{Our best (ViT+UNet, 2-stage)} & 0.924 & 0.804 & 0.754 \\
EVAP                                   & 0.914 & 0.790 & 0.734 \\
SVM                                    & 0.967 & 0.853 & 0.830 \\
\textbf{PCA + SVM}              & \textbf{0.969} & \textbf{0.858} & \textbf{0.835} \\
PCA + KMeans (k=2)                     & 0.439 & 0.559 & 0.257 \\
\addlinespace[2pt]
\midrule
\multicolumn{4}{c}{\textbf{Drought case}}\\
\midrule
Method & PA & UA & IoU \\
\midrule
\textbf{Our best (ViT+UNet, 2-stage)} & \textbf{0.951} & \textbf{0.884} & \textbf{0.845} \\
EVAP                                   & 0.914 & 0.847 & 0.815 \\
SVM                                    & 0.854 & 0.676 & 0.597 \\
PCA + SVM                       & 0.854 & 0.676 & 0.597 \\
PCA + KMeans (k=2)                     & 0.947 & 0.812 & 0.749 \\
\bottomrule
\end{tabular}
\caption{Segmentation performance on two scenarios. Metrics: Producer's Accuracy (PA), User’s Accuracy (UA) , and mean IoU. Classical baselines (SVM / PCA+SVM / PCA+Kmeans) are trained on expert-refined labels; \textbf{EVAP} and \textbf{Ours} do not rely on truth labels.}
\label{tab:cases_pa_ua_iou}
\vspace{2pt}
\end{table}
UA (User Accuracy) reflects precision, PA (Producer Accuracy) reflects recall, and IoU quantifies the overlap between prediction and ground truth.

The classical baselines: Support Vector Machine (SVM) \cite{Cortes1995SVM}, and SVM / Kmeans \cite{MacQueen1967KMeans} on PCA spaces are \emph{supervised}: they are trained on the \emph{manually refined} labels, so higher scores are expected. In contrast, both \textbf{EVAP} and \textbf{Ours} operate \emph{without dense GT}—EVAP applies confidence thresholds, whereas our pipeline expands a small set of expert seeds via PCA-based confidence intervals before segmentation. Hence, the \emph{supervision level differs} across rows in Table \ref{tab:cases_pa_ua_iou}; our claims emphasize label efficiency and operational usability rather than replacing fully supervised upper bounds.

All methods are benchmarked against \emph{expert-refined, full-scene} ground-truth masks provided by TASA, using the same bitemporal inputs, grid, and identical pre/post-processing. We report Intersection-over-Union (IoU), Pixel Accuracy (PA), and User’s Accuracy (UA) on two disaster scenarios.

While \emph{perfect} ground truth would raise absolute scores, such labels are rarely feasible in real disasters due to limited annotation budgets and ambiguous boundaries. Under imperfect references, our shared statistical grounding makes comparisons meaningful: EVAP derives labels via confidence thresholds; our method performs PCA–CI expansion from sparse seeds prior to learning. We treat the TASA masks as the best available proxy and focus on \emph{relative} performance under a consistent protocol. As summarized in Table \ref{tab:cases_pa_ua_iou}, \emph{both} EVAP and our pipeline attain \textbf{comparable PA and UA}, indicating high-quality delineation under a consistent protocol. Our pipeline surpasses EVAP on both cases, highlighting complementary strengths and supporting our emphasis on label-efficient, operational use.

\section{Visualized Results} 
\subsubsection{Label Expansion and Segmentation}
\begin{figure*}[htbp]
    \centering
    \subfloat[Manually annotated polygons, China]{\includegraphics[width=0.48\textwidth]{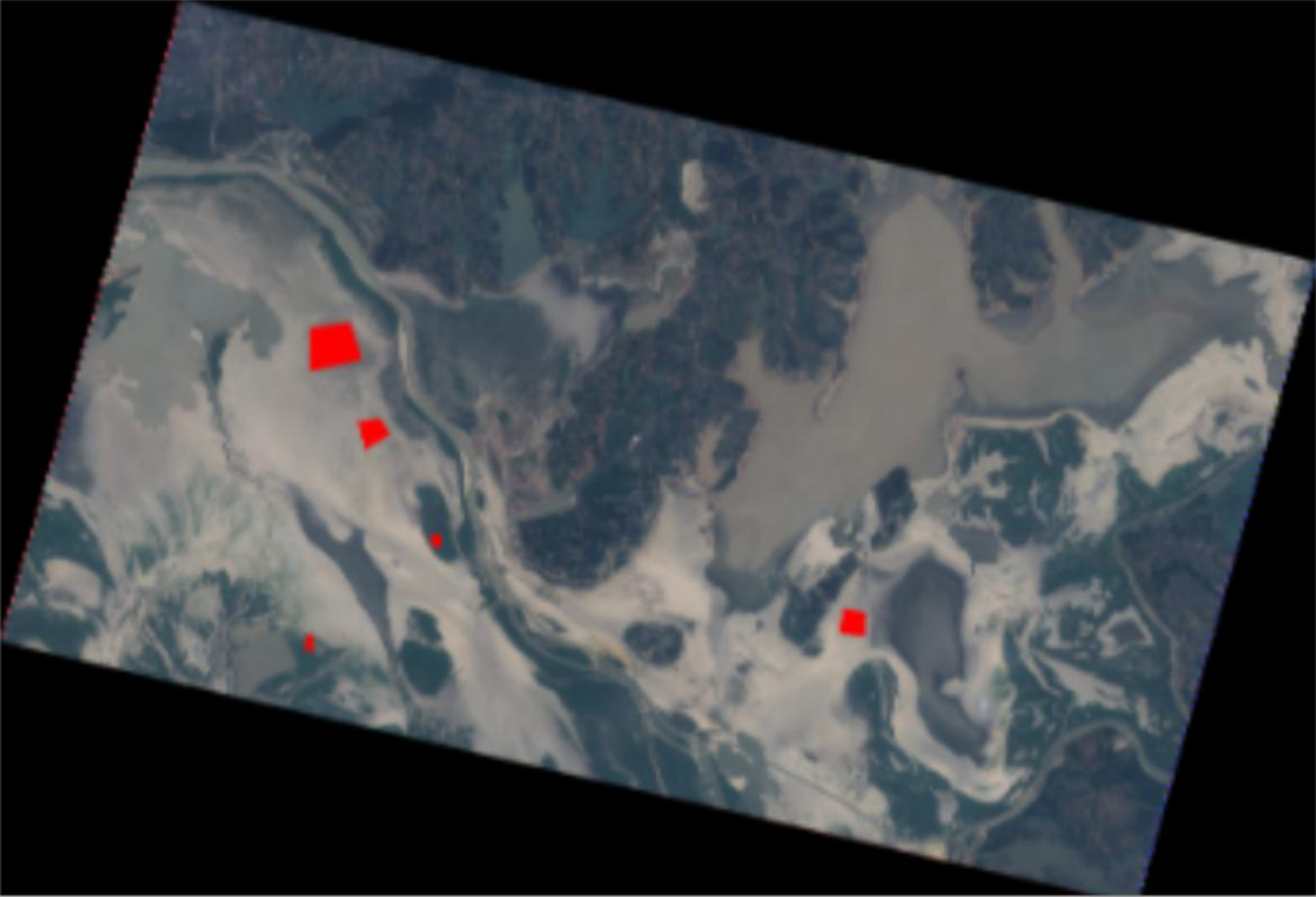}}
    \hfill
    \subfloat[Expanded annotated areas, China]{\includegraphics[width=0.48\textwidth]{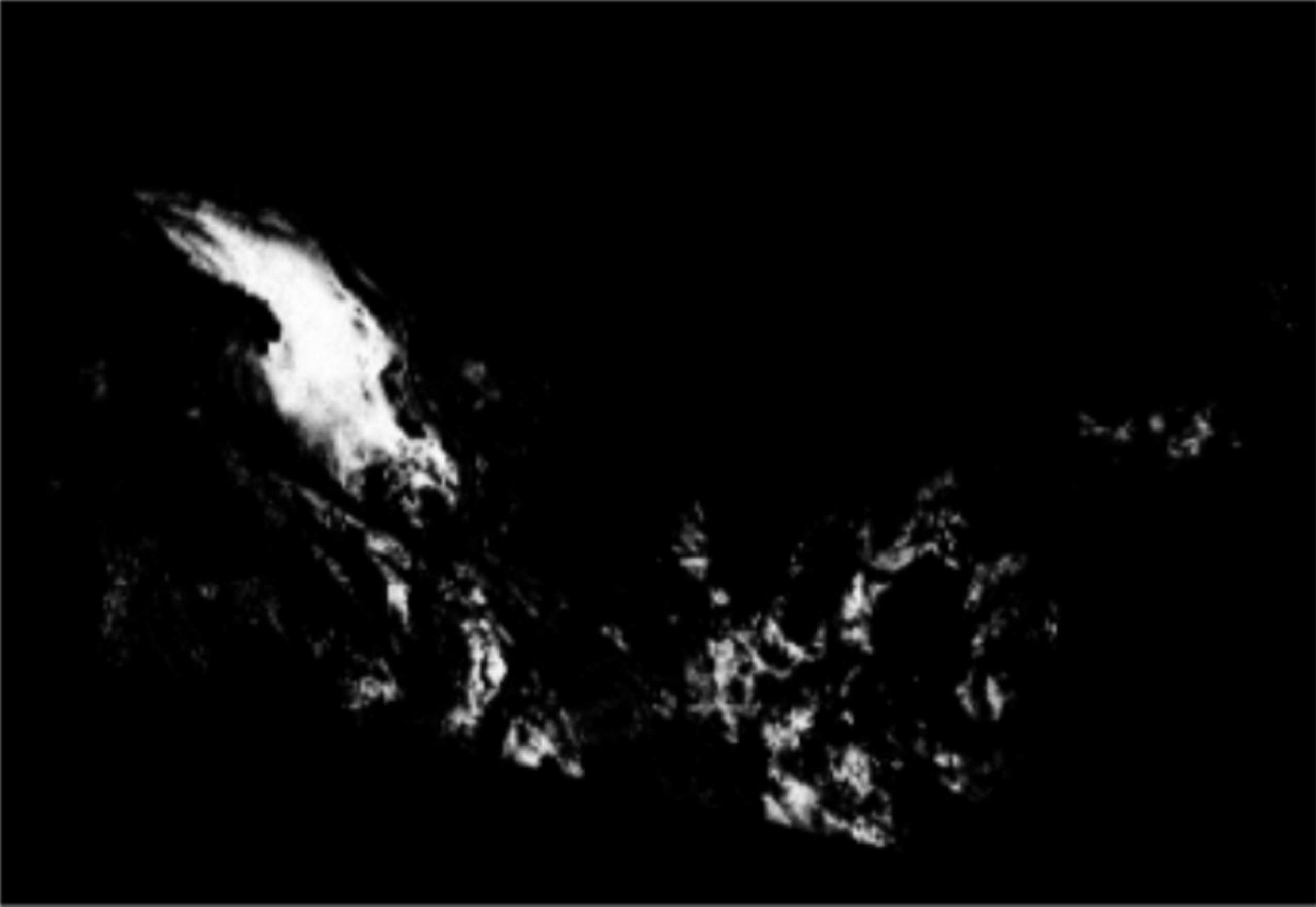}}
    \\
    \subfloat[Manually annotated seed pixels, Greece]{\includegraphics[width=0.48\textwidth]{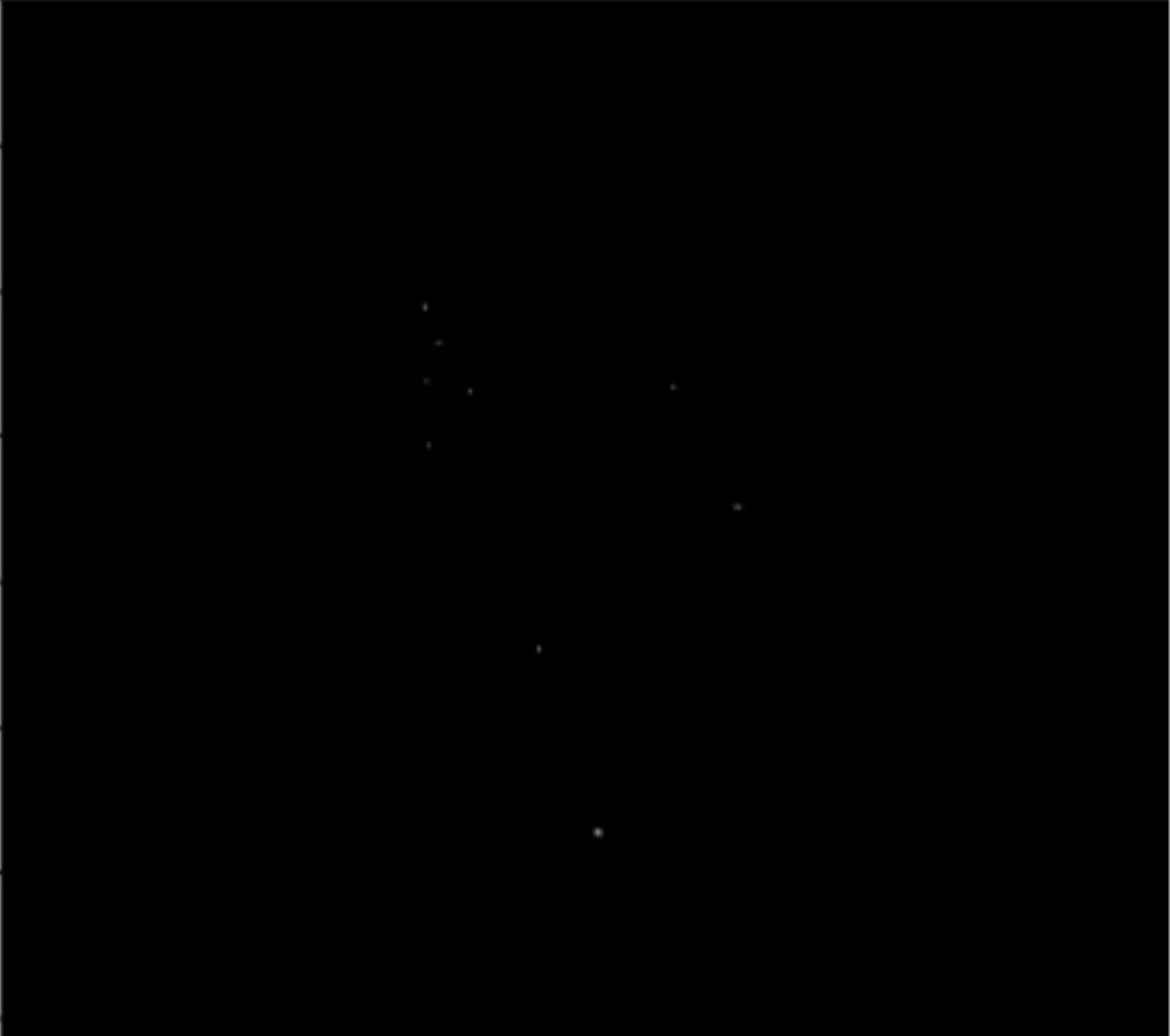}}
    \hfill
    \subfloat[Expanded annotated areas, Greece]{\includegraphics[width=0.48\textwidth]{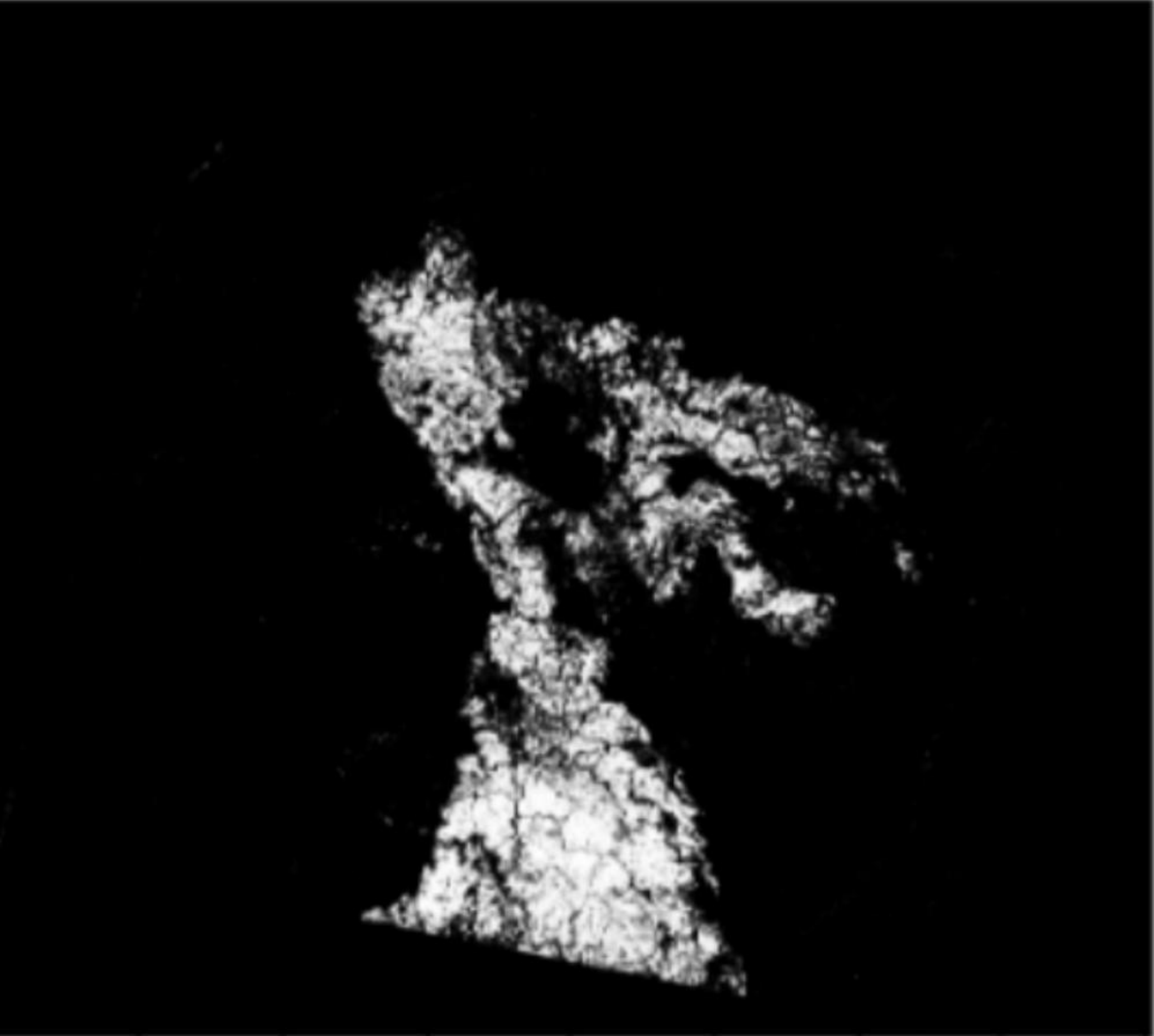}}
    \caption{
    Label initialization and expansion for (a,b) China (Poyang Lake, 2022) and (c,d) Greece (Rhodes, 2023), using PCA-based confidence intervals.
    }
    \label{fig:label_expansion}
\end{figure*}

We first visualize the process and impact of our label expansion strategy. As illustrated in Fig.~\ref{fig:label_expansion}, a small set of manually annotated seed pixels—covering less than 2\% of the image—are projected into PCA space(PC=2) and expanded statistically using a high-confidence Mahalanobis region. This results in substantially enlarged labeled areas, providing dense supervision for subsequent model training in both the Poyang Lake (China) and Rhodes wildfire (Greece) cases.

\begin{figure*}[htbp]
  \centering
  \begin{subfigure}[t]{0.19\textwidth}
    \includegraphics[width=\linewidth]{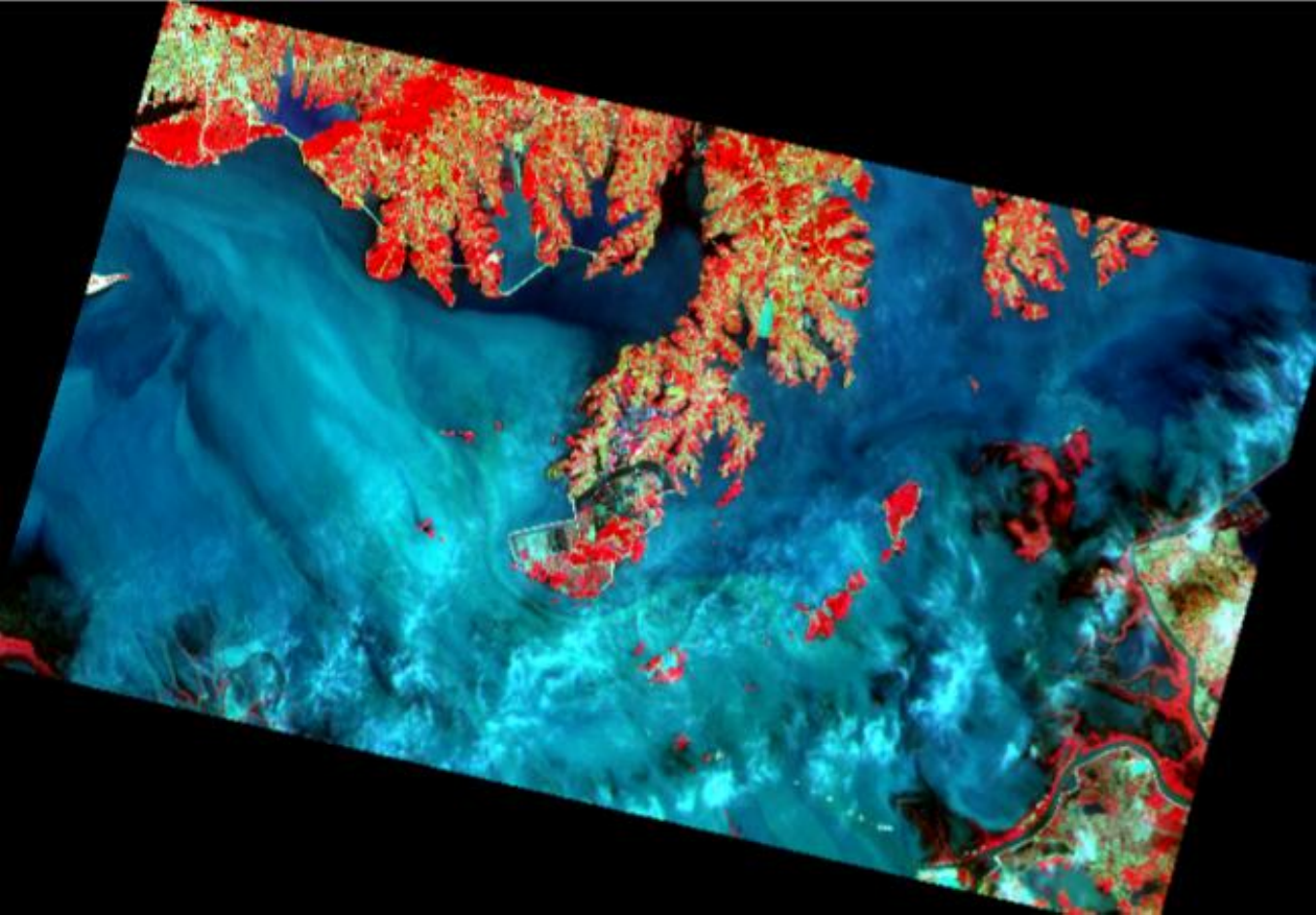}
    \caption{Pre-change (S2)}
    \label{fig:china_pre}
  \end{subfigure}
  \hfill
  \begin{subfigure}[t]{0.19\textwidth}
    \includegraphics[width=\linewidth]{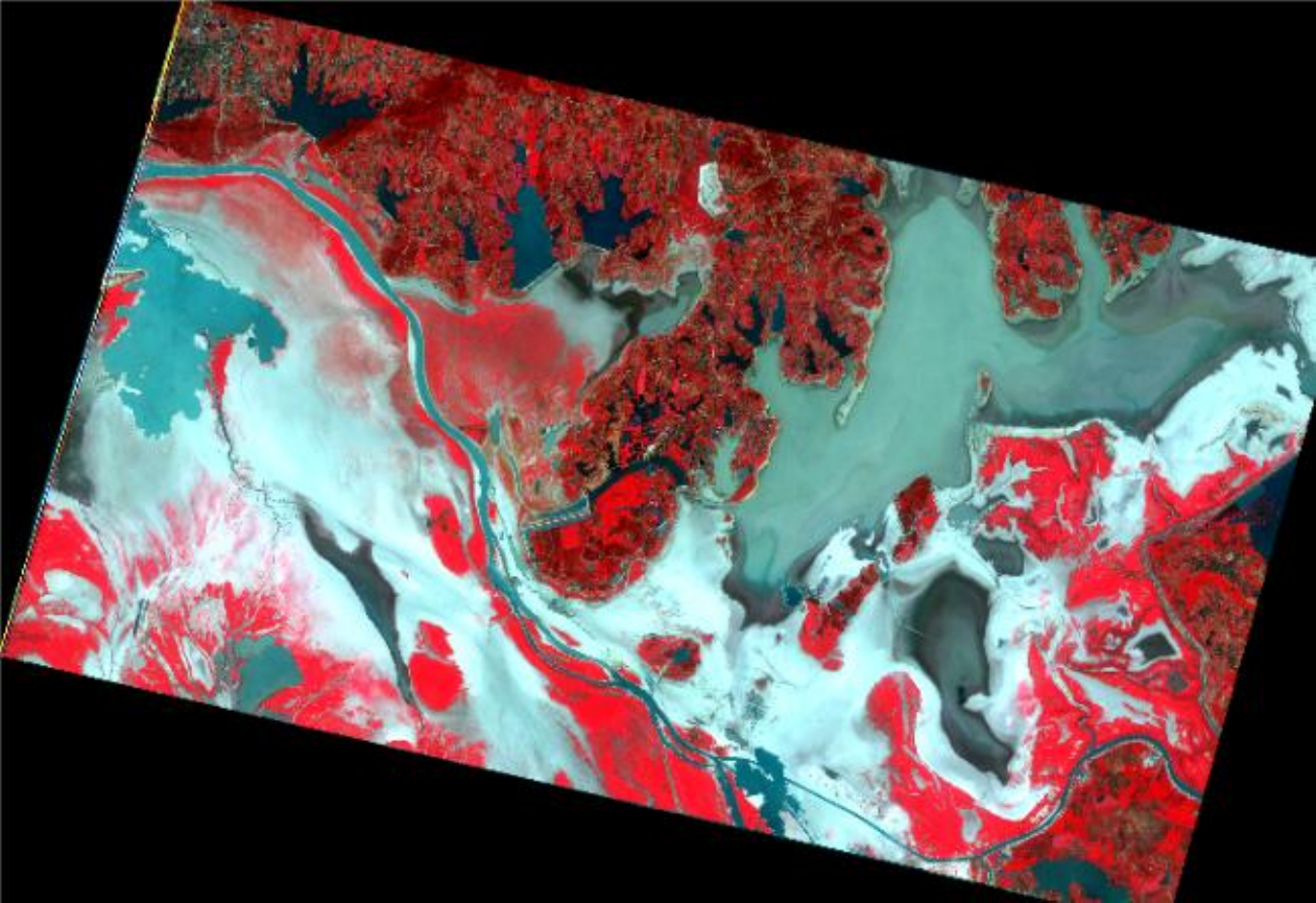}
    \caption{Post-change (FS5)}
    \label{fig:china_post}
  \end{subfigure}
  \hfill
  \begin{subfigure}[t]{0.19\textwidth}
    \includegraphics[width=\linewidth]{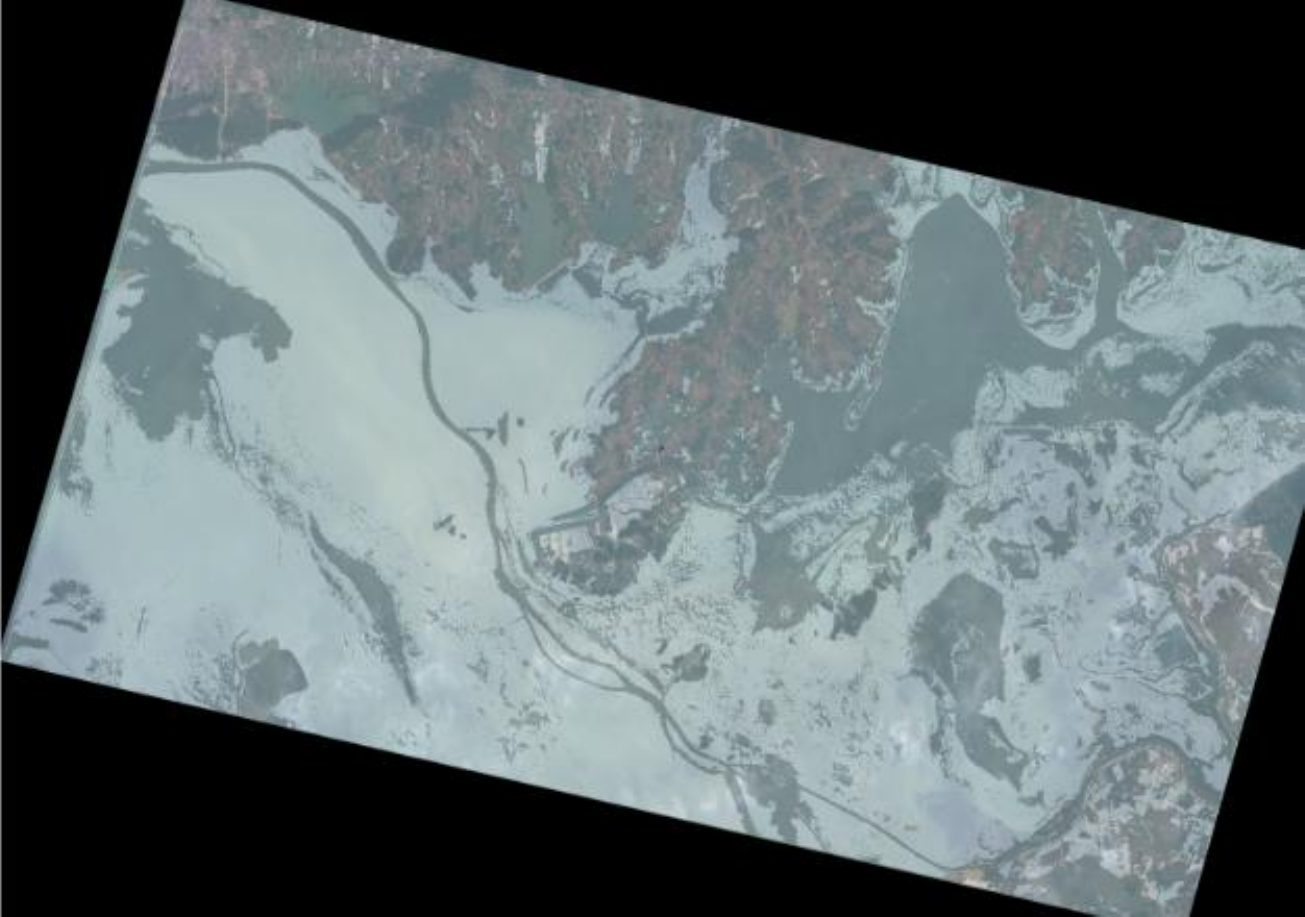}
    \caption{EVAP result}
    \label{fig:china_evap}
  \end{subfigure}
  \hfill
  \begin{subfigure}[t]{0.19\textwidth}
    \includegraphics[width=\linewidth]{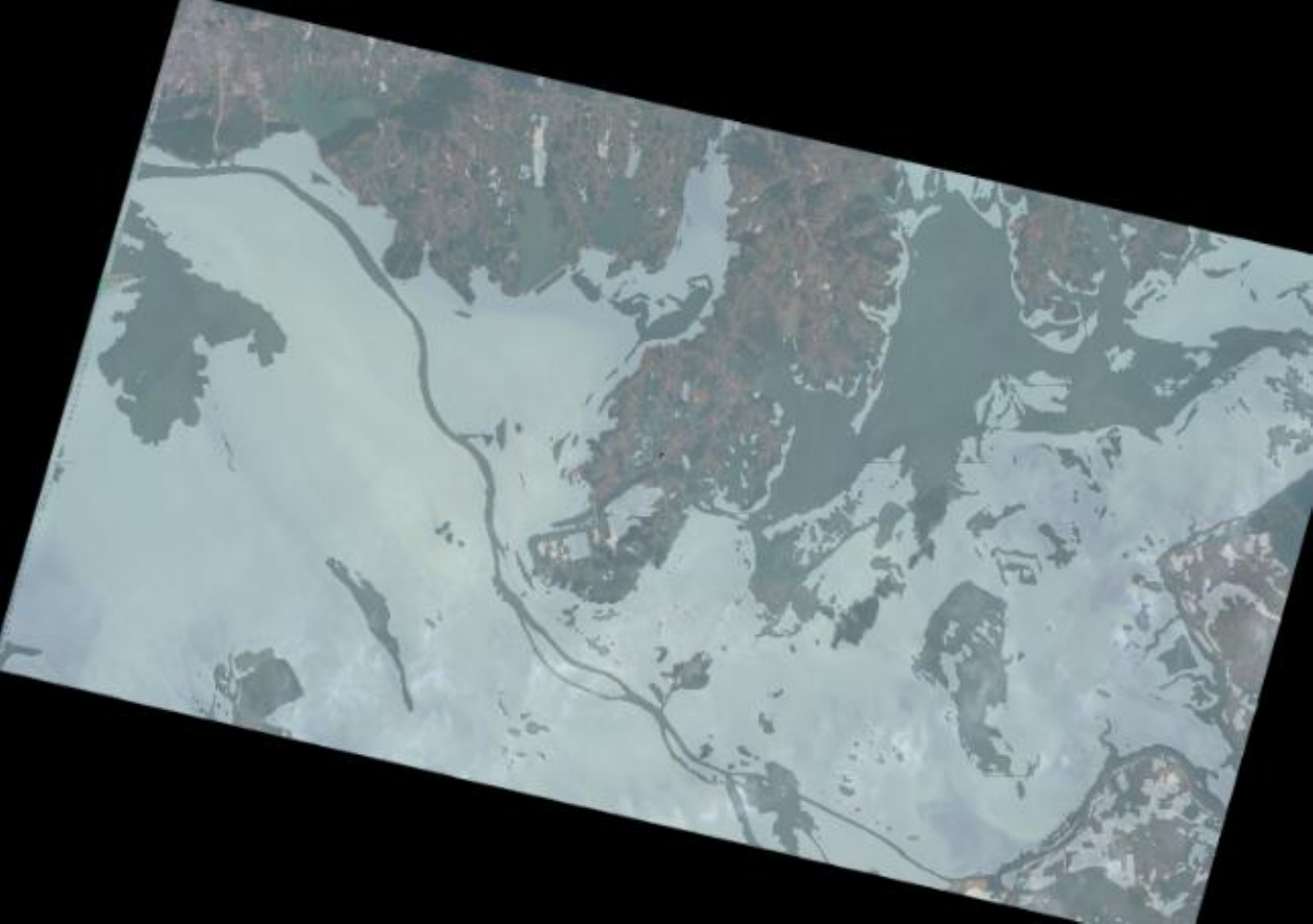}
    \caption{Our model}
    \label{fig:china_our}
  \end{subfigure}
  \hfill
  \begin{subfigure}[t]{0.19\textwidth}
    \includegraphics[width=\linewidth]{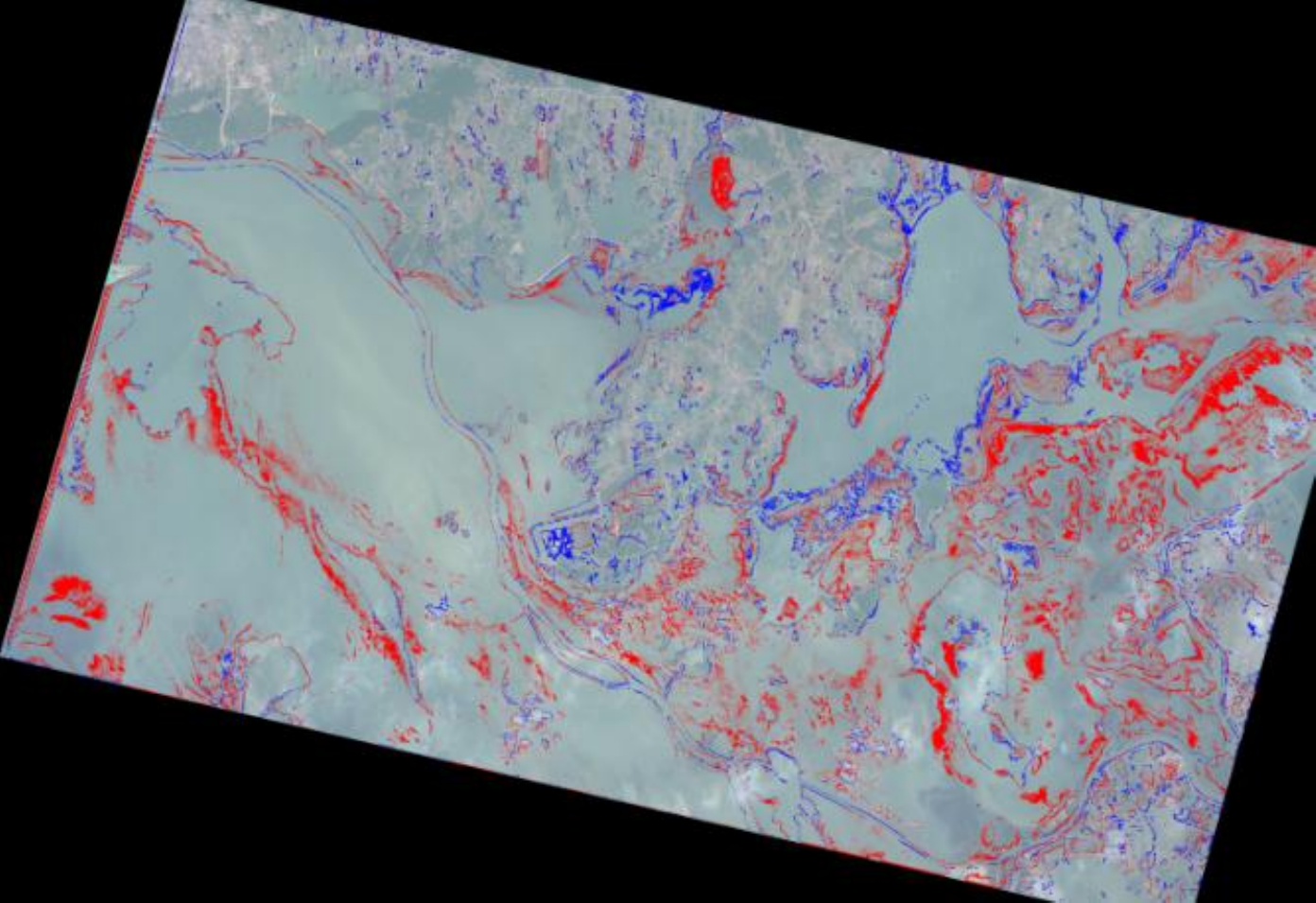}
    \caption{Difference map}
    \label{fig:china_diff}
  \end{subfigure}
  \caption{Segmentation results of the 2022 Poyang Lake drought event in China. The images show: (a) pre-change false-color image from Sentinel-2 (S2), (b) post-change false-color image from Formosat-5 (FS5), (c) EVAP output on natural color image, (d) our model's prediction on natural color image, and (e) their pixel-wise difference on natural color image. In the difference map, gray indicates predicted affected area, red marks commission errors (false positives), and blue denotes omission errors (false negatives).}
  \label{fig:china_segmentation}
\end{figure*}

\begin{figure*}[htbp]
  \centering
  \begin{subfigure}[t]{0.19\textwidth}
    \includegraphics[width=\linewidth]{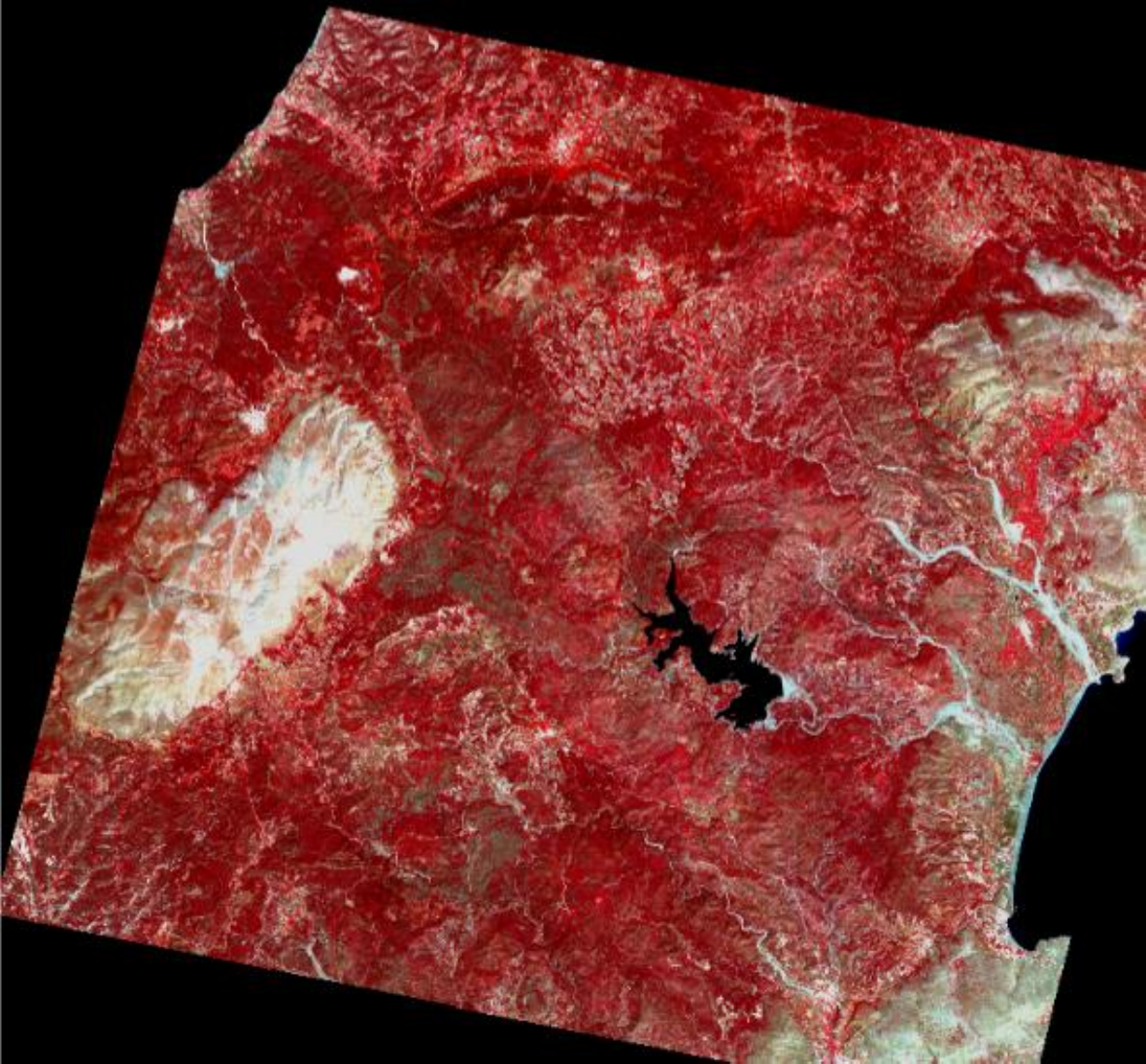}
    \caption{Pre-change (S2)}
    \label{fig:greek_pre}
  \end{subfigure}
  \hfill
  \begin{subfigure}[t]{0.19\textwidth}
    \includegraphics[width=\linewidth]{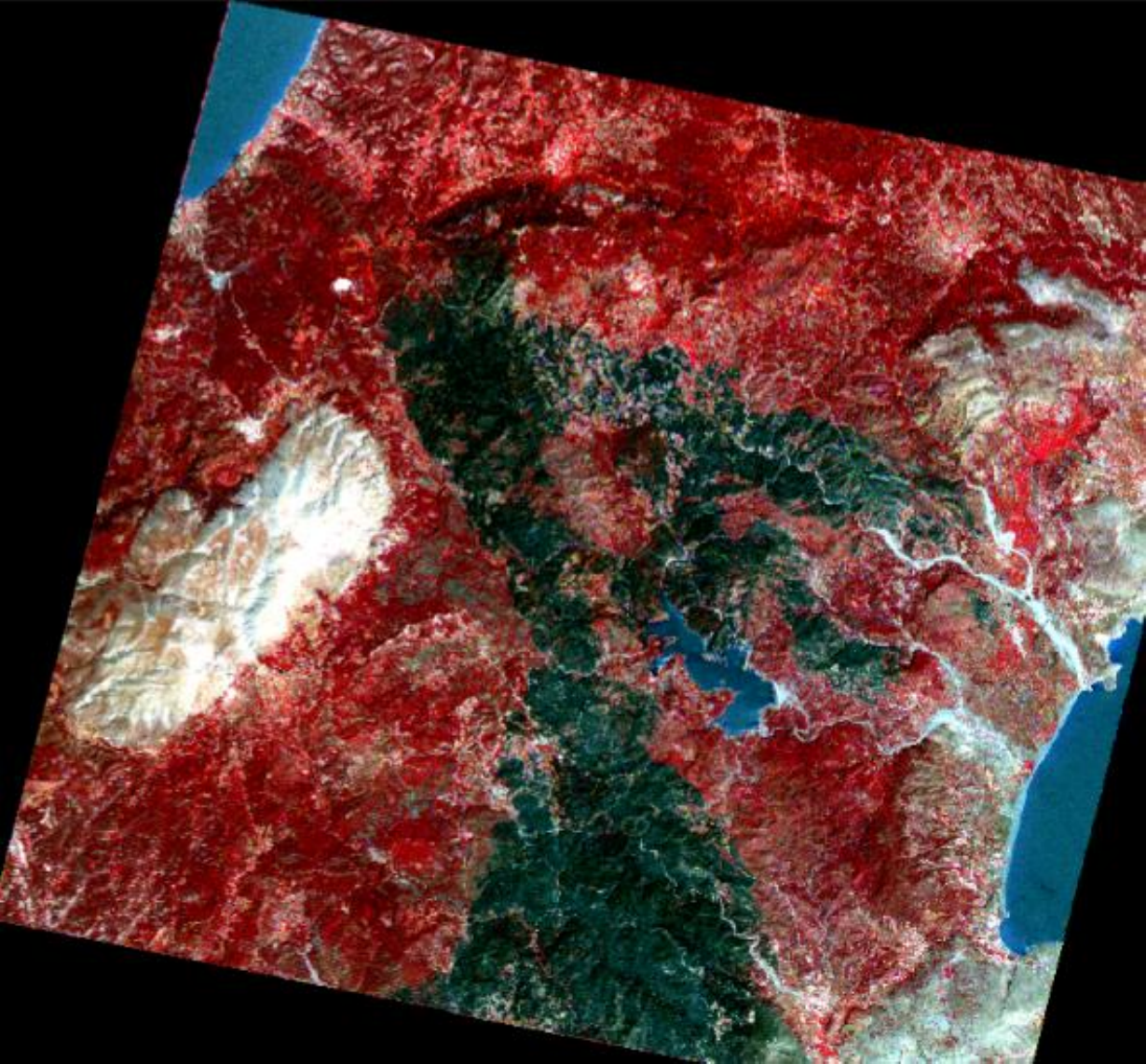}
    \caption{Post-change (FS5)}
    \label{fig:greek_post}
  \end{subfigure}
  \hfill
  \begin{subfigure}[t]{0.19\textwidth}
    \includegraphics[width=\linewidth]{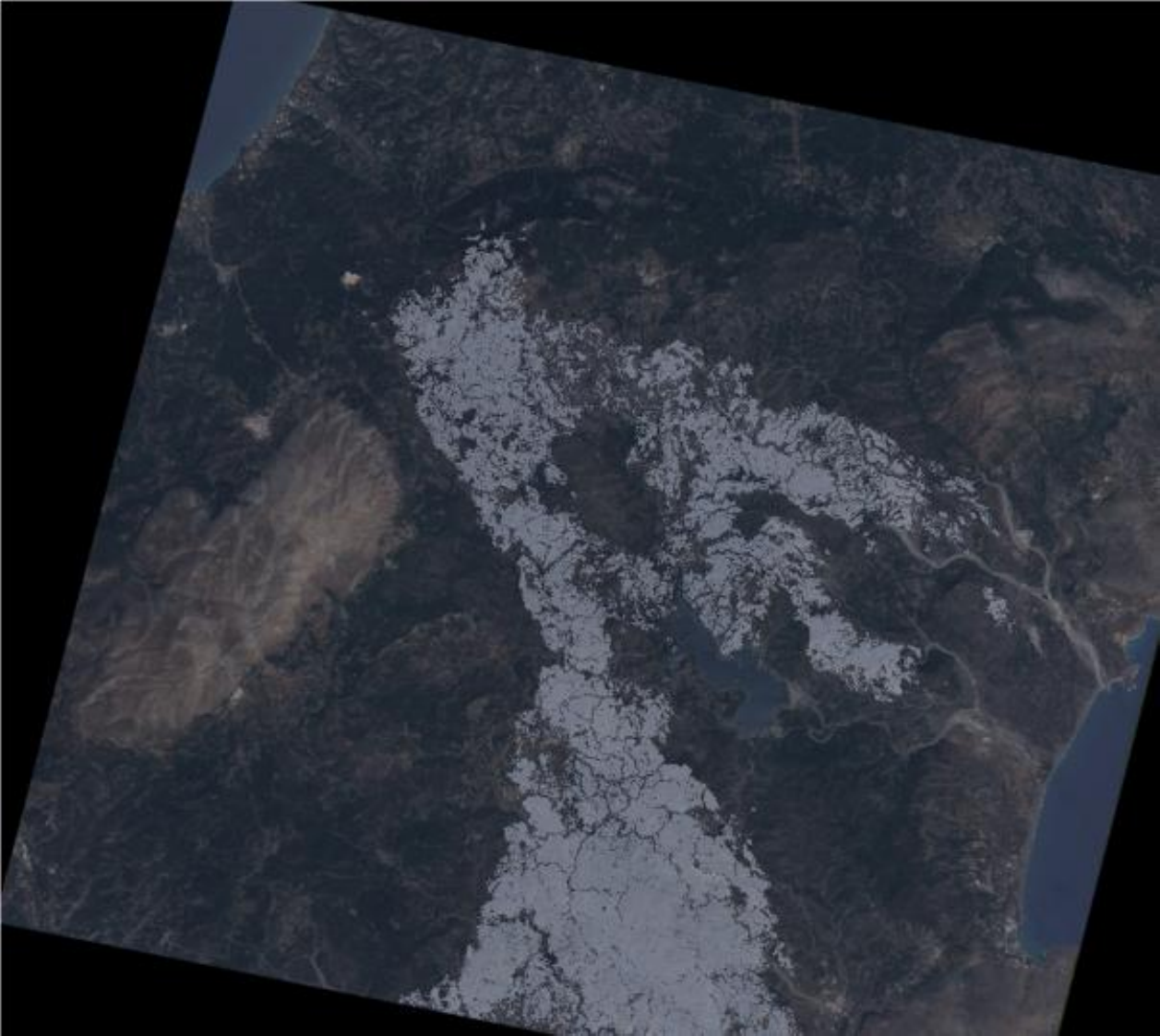}
    \caption{EVAP result}
    \label{fig:greek_evap}
  \end{subfigure}
  \hfill
  \begin{subfigure}[t]{0.19\textwidth}
    \includegraphics[width=\linewidth]{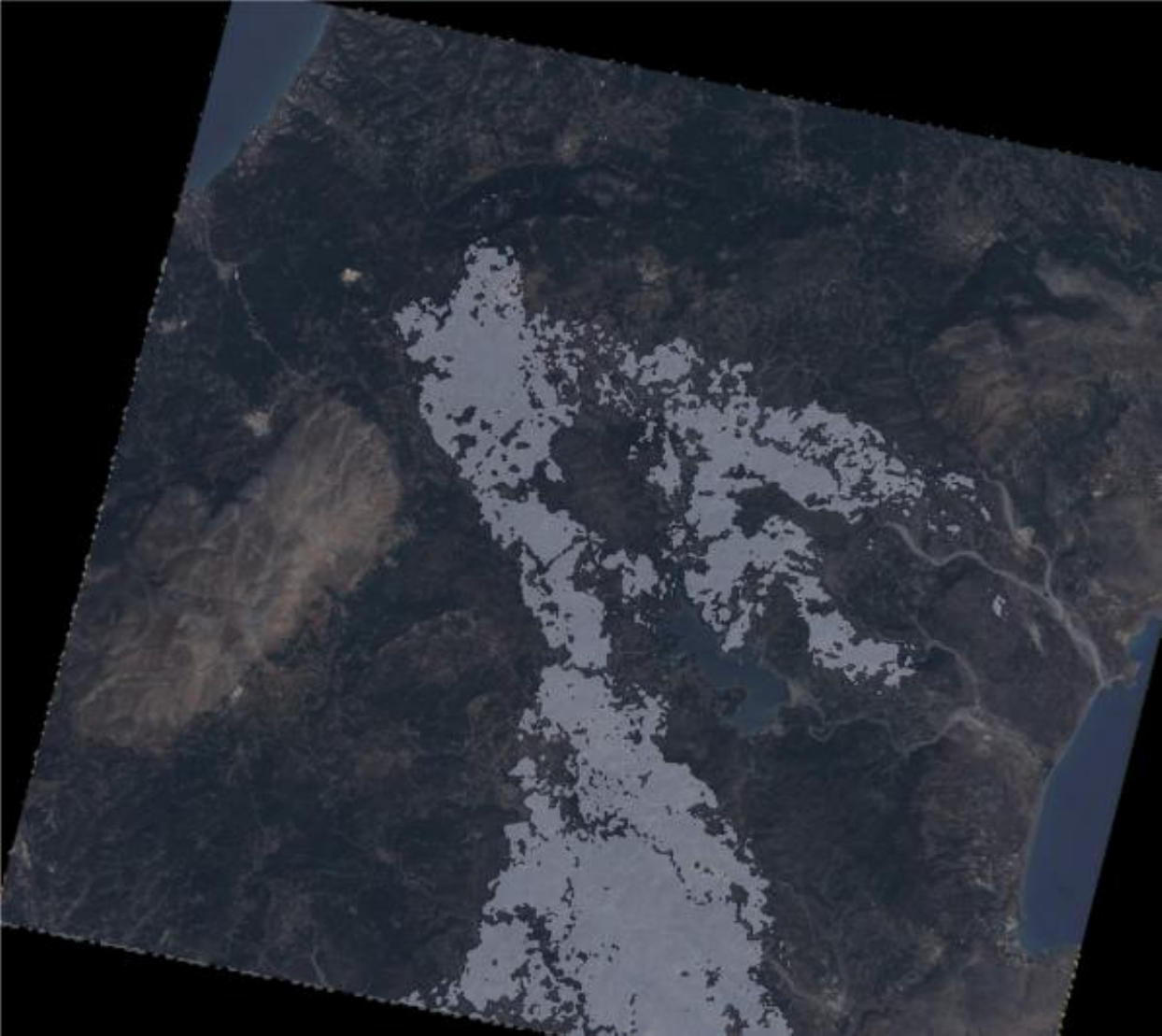}
    \caption{Our model}
    \label{fig:greek_our}
  \end{subfigure}
  \hfill
  \begin{subfigure}[t]{0.19\textwidth}
    \includegraphics[width=\linewidth]{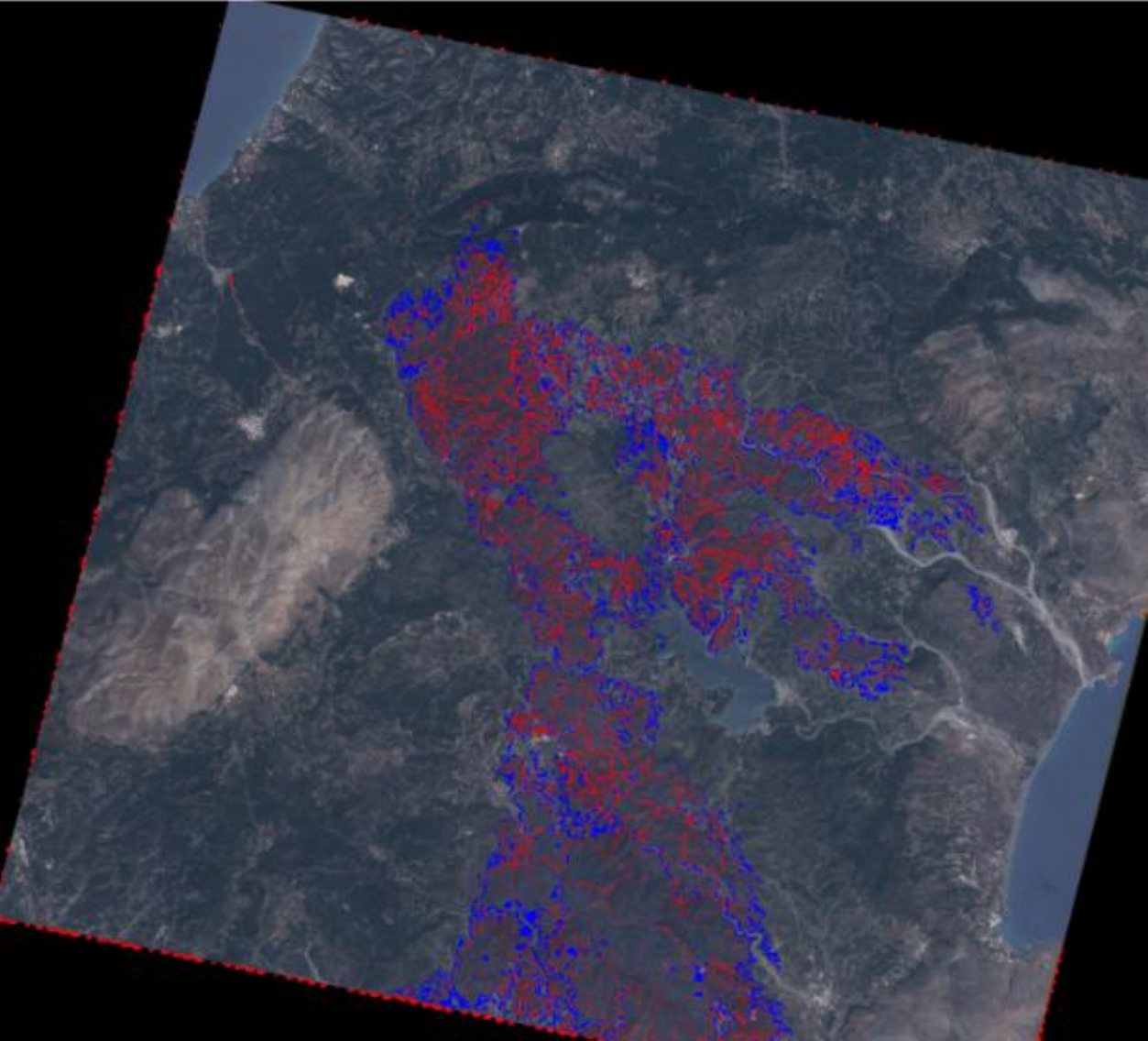}
    \caption{Difference map}
    \label{fig:greek_diff}
  \end{subfigure}
  \caption{Segmentation results of the 2023 Rhodes wildfire event in Greece. The images show: (a) pre-change false-color image from Sentinel-2 (S2), (b) post-change false-color image from Formosat-5 (FS5), (c) EVAP output mask on natural color image, (d) our model's prediction on natural color image, and (e) their pixel-wise difference on natural color image. In the difference map, gray indicates predicted affected area, red marks commission errors (false positives), and blue denotes omission errors (false negatives).}
  \label{fig:greece_segmentation}
\end{figure*}

Fig.~\ref{fig:china_segmentation} and Fig.~\ref{fig:greece_segmentation} present whole-scene segmentation results for both study areas. For each event, we compare (1) pre- and post-event imagery, (2) the baseline EVAP segmentation, (3) our model’s prediction, and (4) a difference map highlighting commission (red) and omission (blue) errors. For the Greece wildfire case, the model output is generated using decoder A with the BCE loss, while for the Poyang Lake drought case, results are obtained from the model with decoder C and the two-stage loss strategy. In both cases, our model more accurately delineates the affected regions and reduces both types of errors, indicating superior generalization over the baseline.

\subsubsection{Zoom-in Comparison and Boundary Smoothness}
\begin{figure*}[!t]
    \centering
    \subfloat[EVAP, China]{\includegraphics[width=0.48\textwidth]{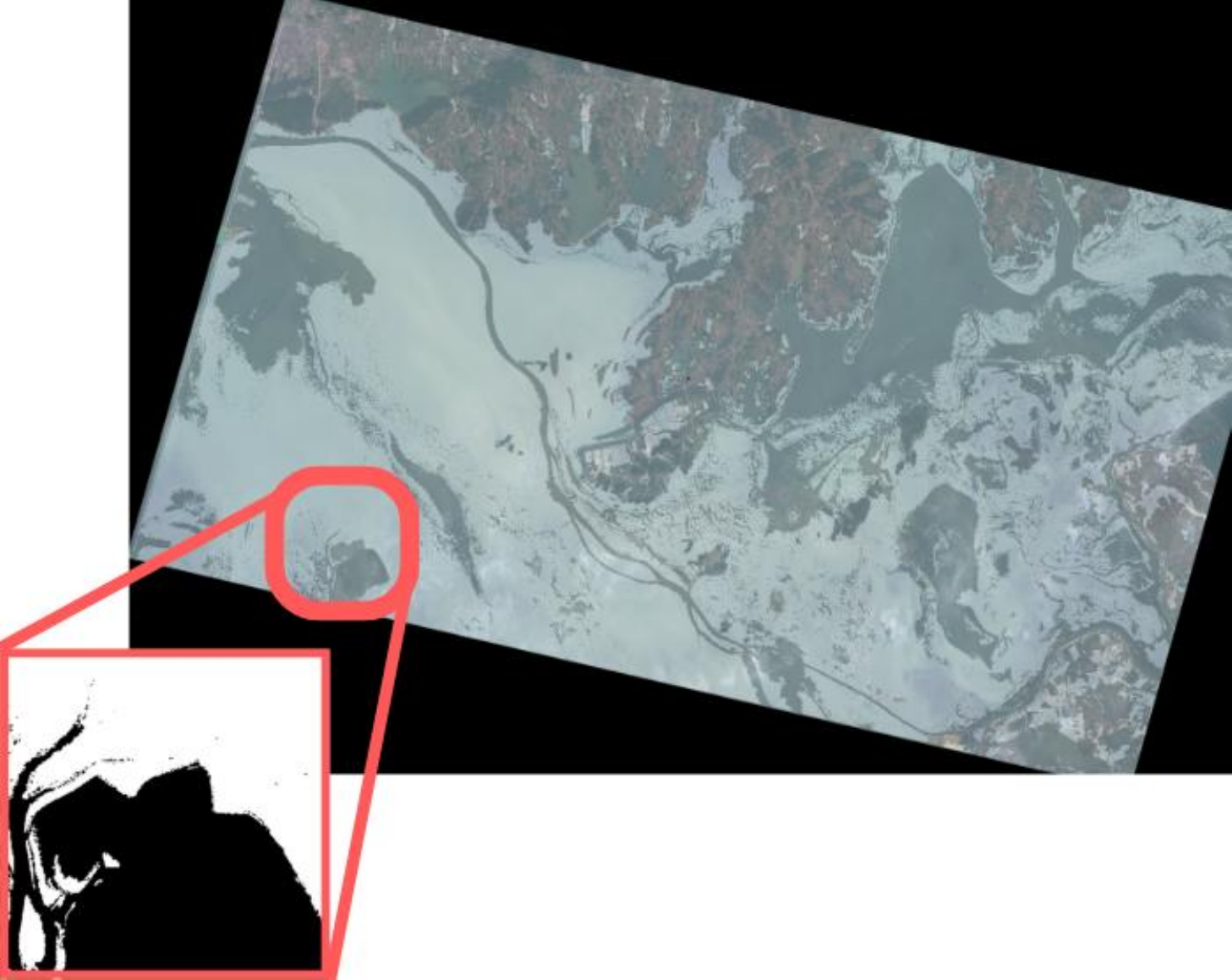}}
    \hfill
    \subfloat[Ours, China]{\includegraphics[width=0.48\textwidth]{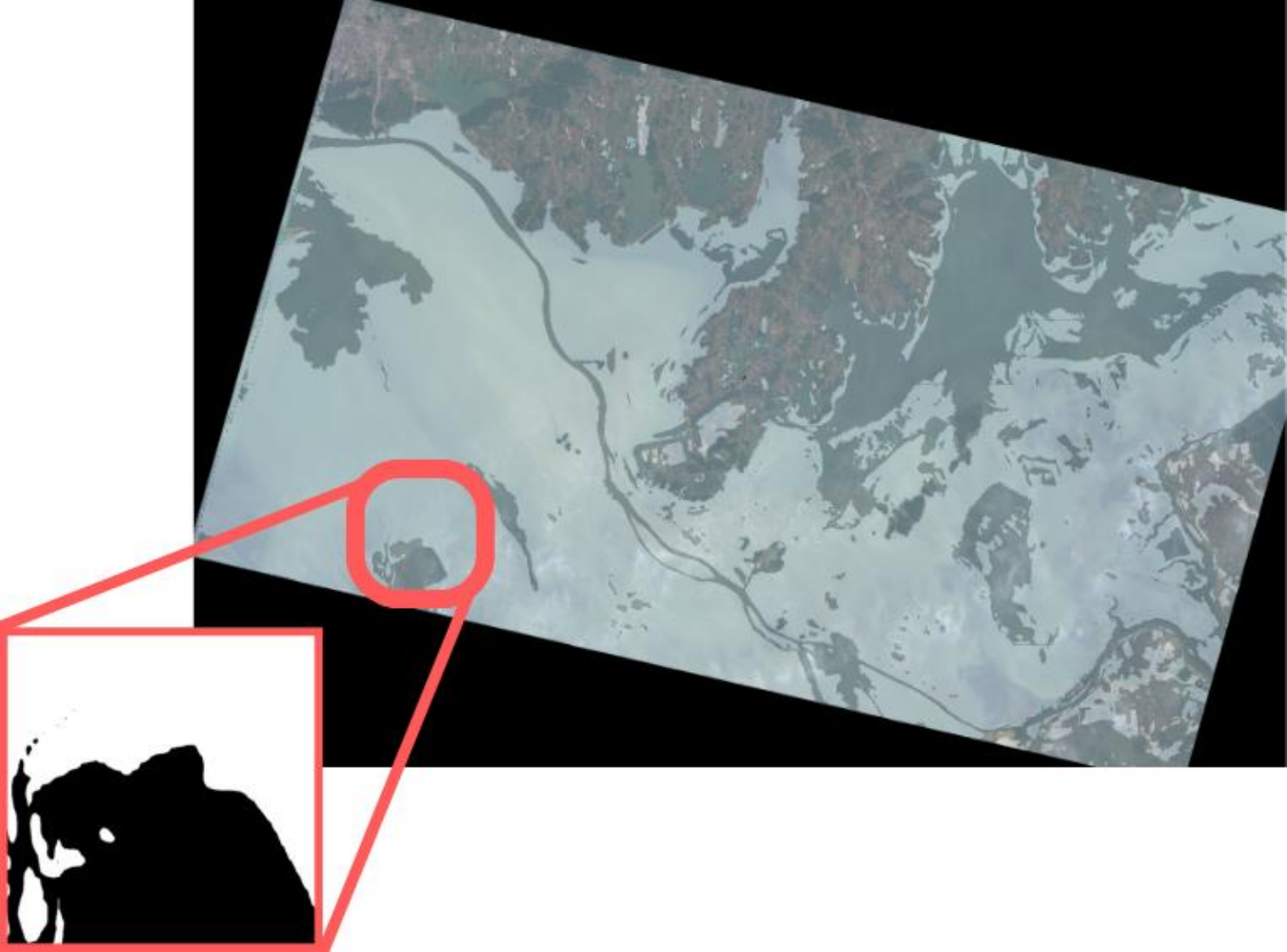}}
    \\
    \subfloat[EVAP, Greece]{\includegraphics[width=0.48\textwidth]{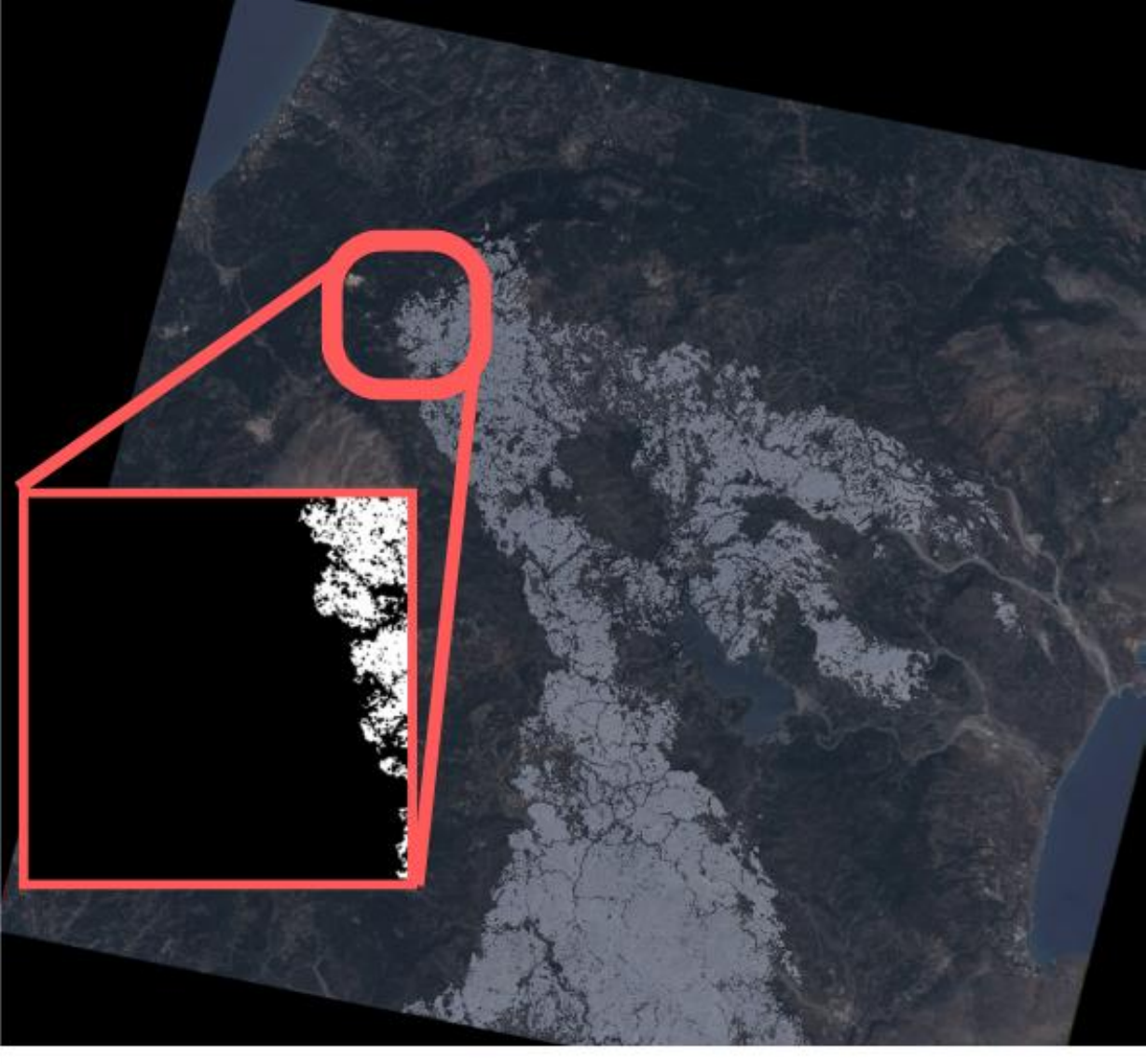}}
    \hfill
    \subfloat[Ours, Greece]{\includegraphics[width=0.48\textwidth]{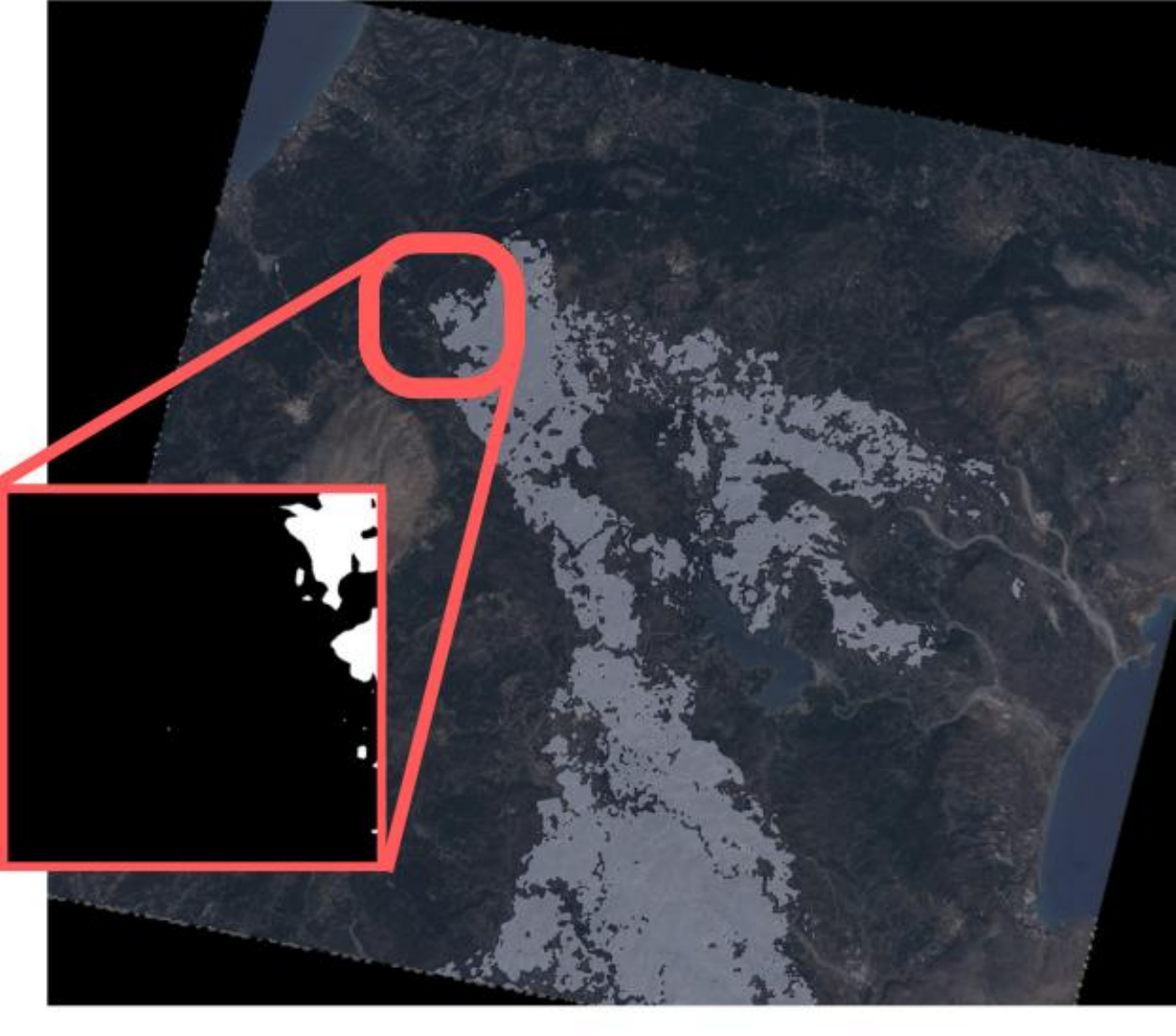}}
    \caption{
    Close-up comparison of segmentation results from EVAP and our model for China (Poyang Lake drought, 2022) and Greece (Rhodes wildfire, 2023), highlighting differences in boundary accuracy.
    }
    \label{fig:zoomed_comparison}
\end{figure*}

To further investigate segmentation quality, we present zoomed-in comparisons of representative regions in Fig.~\ref{fig:zoomed_comparison}. It is evident that the outputs of our model are notably smoother and less fragmented than those produced by EVAP. In the context of natural disaster mapping, such as wildfire and drought, contiguous affected areas are more plausible than highly fragmented patches and sparse pixels. The improved smoothness of the boundary and spatial coherence of the predictions of our model suggest that our method provides a closer approximation to the true extent of disaster-affected regions, even in the absence of a perfect ground truth.

These qualitative results complement our quantitative findings, underscoring the advantage of combining data-driven label expansion with transformer-based segmentation for robust and realistic disaster mapping.

\section{Conclusion}
In this work, we propose a robust, semi-automatic framework for disaster-affected area segmentation using multi-satellite imagery. By integrating PCA-based label expansion and transformer-based deep learning architectures, our method effectively addresses the challenge of limited manual annotations and achieves superior segmentation performance compared to the baseline EVAP system. Both quantitative and qualitative results on real-world wildfire and drought scenarios demonstrate that our approach consistently improves the delineation of affected regions and produces spatially coherent segmentation maps. Vision transformer(ViT)-based models, in particular, exhibit notable stability and rapid convergence, further highlighting their suitability for operational disaster response tasks.

The proposed pipeline is designed to be both sensor- and hazard-agnostic, exhibiting encouraging transferability across representative sub-regions. Generalization may nonetheless be influenced by spectral/resolution mismatches between sensors, seasonal shifts, and cloud/shadow conditions. Importantly, the framework requires manual annotation only for a representative subset of the affected area and can automatically classify analogous regions. In cross-sensor experiments, the approach maintained comparable performance.

For practical use in real disasters, a minimal deployment recipe is: (i) tile incoming scenes with small overlap and run quick image quality checks; (ii) derive a small set of manually labeled polygons and apply our PCA-based expansion to obtain pseudo-labels for training; (iii) run a ViT encoder with a lightweight decoder for per-tile inference; and (iv) apply simple morphology, stitch tiles back to the scene grid, and export. This keeps integration straightforward while preserving the label-efficiency benefits demonstrated in our experiments.

Although our current framework has shown effectiveness and stability in emerging disaster analysis, several avenues remain for future work. Potential directions include the incorporation of active learning strategies to further minimize manual labeling effort, the extension and experiments of this method to additional disaster types, and the integration of additional data sources (e.g., SAR,  or meteorological data) to improve model generalization. In addition, future research could explore the implementation and deployment in real time within operational emergency response systems. In general, our results provide a promising foundation for advancing automated disaster mapping in remote sensing applications.

\section*{Acknowledgments}
The authors thank the Taiwan Space Agency (TASA) for providing technical support and access to satellite datasets. Special thanks are extended to Dr. Li-Yu Chang, Director of the Division of Satellite Data Applications at TASA, for his valuable guidance and support. The authors also acknowledge helpful discussions with colleagues at TASA.

During the preparation of this manuscript, the authors used ChatGPT (OpenAI, GPT-4o, 2024) for purposes of English language refinement, scientific editing, and technical clarifications. The authors have reviewed and edited all outputs, and take full responsibility for the content of this publication.

\subsection*{Abbreviations}{
The following abbreviations are used in this manuscript:
\\
\noindent 
\begin{tabular}{@{}ll}
AI       & Artificial Intelligence \\
B        & Blue \\
CAM      & Class Activation Mapping \\
CI       & Confidence Interval \\
CNN      & Convolutional Neural Network \\
CV       & Computer Vision \\
EVAP     & Emergent Value Added Product \\
FS5      & Formosat-5 \\
FTN      & Fully Transformer Network \\
G        & Green \\
GIS      & Geographic Information System \\
IoU      & Intersection over Union \\
MLP      & Multi-Layer Perceptron \\
MSI      & MultiSpectral Instructment\\
NBR      & Normalized Burn Ratio \\
NDVI     & Normalized Difference Vegetation Index \\
NDWI     & Normalized Difference Water Index \\
NIR      & Near-Infrared \\
PAN      & Panchromatic \\
MS       & Multispectral \\
PCA      & Principal Component Analysis \\
R        & Red \\
S2       & Sentinel-2 \\
SAM      & Segment Anything Model \\
SAR      & Synthetic Aperture Radar \\
TASA     & Taiwan Space Agency \\
VAP      & Value Added Product \\
VHR      & Very High Resolution \\
ViT      & Vision Transformer \\
\end{tabular}
}

\bibliographystyle{plainnat}
\bibliography{ref}

@inproceedings{hung2024evap,
  author       = {Jung-Chien Hung and Li-Yu Chang},
  title        = {Emergent Value-Added Product Processing by Gaussian Statistical Approach for Sentinel-2 Data},
  booktitle    = {ASGC 2024 Conference},
  year         = {2024},
  organization = {Taiwan Space Agency},
  address      = {Hsinchu, Taiwan},
  url          = {https://indico4.twgrid.org/event/33/contributions/1454/attachments/784/986/ASGC2024_EVAP_Taiwan%20Space%20Agency.pdf},
  note         = {Accessed: 2025-05-15}
}

@article{dosovitskiy2020vit,
  title        = {An Image is Worth 16x16 Words: Transformers for Image Recognition at Scale},
  author       = {Dosovitskiy, Alexey and Beyer, Lucas and Kolesnikov, Alexander and Weissenborn, Dirk and Zhai, Xiaohua and Unterthiner, Thomas and Dehghani, Mostafa and Minderer, Matthias and Heigold, Georg and Gelly, Sylvain and Uszkoreit, Jakob and Houlsby, Neil},
  journal      = {arXiv preprint arXiv:2010.11929},
  year         = {2020},
  url          = {https://arxiv.org/abs/2010.11929},
  doi          = {10.48550/arXiv.2010.11929}
}

@inproceedings{bandara2022changeformer,
  author    = {Bandara, Wele Gedara Chaminda and Patel, Vishal M.},
  title     = {A Transformer-Based Siamese Network for Change Detection},
  booktitle = {IGARSS 2022 -- IEEE International Geoscience and Remote Sensing Symposium},
  year      = {2022},
  pages     = {207--210},
  doi       = {10.1109/IGARSS46834.2022.9883686}
}

@article{chen2020levircd,
  title   = {LEVIR-CD: A High-Resolution Remote Sensing Dataset for Object-Level Change Detection},
  author  = {Chen, Hao and Shi, Zhenwei},
  journal = {arXiv preprint arXiv:2003.07756},
  year    = {2020},
  url     = {https://arxiv.org/abs/2003.07756}
}

@inproceedings{gupta2019xbd,
  title     = {Creating xBD: A Dataset for Assessing Building Damage from Satellite Imagery},
  author    = {Gupta, Ritwik and Hosfelt, Russell and Sajeev, Sachit and Heim, Eric and Doshi, Jigar and Lucas, Keane and Choset, Howie and Gaston, Matthew E.},
  booktitle = {Proceedings of the IEEE/CVF Conference on Computer Vision and Pattern Recognition Workshops (CVPRW)},
  year      = {2019},
  pages     = {10--17},
  url       = {https://openaccess.thecvf.com/content_CVPRW_2019/html/WAD/Gupta_Creating_xBD_A_Dataset_for_Assessing_Building_Damage_from_Satellite_CVPRW_2019_paper.html}
}

@misc{xview2challenge,
  title        = {xView2 Challenge: Assessing Building Damage from Satellite Imagery},
  author       = {{Defense Innovation Unit (DIU)} and {Carnegie Mellon University Software Engineering Institute (SEI)}},
  year         = {2019},
  howpublished = {\url{https://xview2.org/}},
  note         = {Accessed: 2025-05-15}
}

@article{chen2017rethinking,
  title   = {Rethinking Atrous Convolution for Semantic Image Segmentation},
  author  = {Chen, Liang-Chieh and Papandreou, George and Schroff, Florian and Adam, Hartwig},
  journal = {arXiv preprint arXiv:1706.05587},
  year    = {2017},
  url     = {https://arxiv.org/abs/1706.05587}
}

@article{fakhri2025flood,
  title   = {Quantitative evaluation of flood extent detection using attention U-Net: case studies from Eastern South Wales, Australia in March 2021 and July 2022},
  author  = {Fakhri, A. and Gkanatsios, K.},
  journal = {Scientific Reports},
  year    = {2025},
  volume  = {15},
  pages   = {92734},
  doi     = {10.1038/s41598-025-92734-x}
}

@article{khankeshizadeh2024,
  title   = {FBA-DPAttResU-Net: Forest burned area detection using a novel end-to-end dual-path attention residual-based U-Net from post-fire Sentinel-1 and Sentinel-2 images},
  author  = {Khankeshizadeh, E. and Smith, J. and Lee, H.},
  journal = {Remote Sensing},
  volume  = {16},
  number  = {8},
  pages   = {1456},
  year    = {2024},
  doi     = {10.3390/rs16081456}
}

@article{chen2020spatial,
  title   = {A Spatial-Temporal Attention-Based Method and a New Dataset for Remote Sensing Image Change Detection},
  author  = {Chen, Hao and Shi, Zhenwei},
  journal = {Remote Sensing},
  volume  = {12},
  number  = {10},
  pages   = {1662},
  year    = {2020},
  doi     = {10.3390/rs12101662}
}

@article{wang2022unetformer,
  author  = {Wang, Libo and Li, Rui and Zhang, Ce and Fang, Shenghui and Duan, Chenxi and Meng, Xiaoliang and Atkinson, Peter M.},
  title   = {UNetFormer: A U-Net-like Transformer for efficient semantic segmentation of remote sensing urban scene imagery},
  journal = {ISPRS Journal of Photogrammetry and Remote Sensing},
  volume  = {190},
  pages   = {196--214},
  year    = {2022},
  doi     = {10.1016/j.isprsjprs.2022.06.008}
}

@inproceedings{Yan2022FTN,
  author    = {Yan, Tianyu and Wan, Zifu and Zhang, Pingping},
  title     = {Fully Transformer Network for Change Detection of Remote Sensing Images},
  booktitle = {Computer Vision -- ACCV 2022, Part II (Lecture Notes in Computer Science, vol. 13676)},
  pages     = {75--92},
  publisher = {Springer},
  year      = {2023},
  doi       = {10.1007/978-3-031-26284-5_5}
}

@article{Cao2023Weak,
  author  = {Cao, Yinxia and Huang, Xiaoqin},
  title   = {A coarse-to-fine weakly supervised learning method for green plastic cover segmentation using high-resolution remote sensing images},
  journal = {ISPRS Journal of Photogrammetry and Remote Sensing},
  volume  = {188},
  pages   = {157--176},
  year    = {2022},
  doi     = {10.1016/j.isprsjprs.2022.04.024}
}

@article{lu2024weak,
  author  = {Lu, Xiao and Jiang, Zhiguo and Zhang, Haopeng},
  title   = {Weakly Supervised Remote Sensing Image Semantic Segmentation with Pseudo-Label Noise Suppression},
  journal = {IEEE Transactions on Geoscience and Remote Sensing},
  volume  = {62},
  year    = {2024},
  doi     = {10.1109/TGRS.2023.3290378}
}

@article{kirillov2023sam,
  author  = {Kirillov, Alexander and Mintun, Eric and Ravi, Nikhila and Mao, Hanzi and Rolland, Chloe and Gustafson, Laura and Xiao, Tete and Whitehead, Spencer and Berg, Alexander C. and Lo, Wan-Yen and Doll{\'a}r, Piotr and Girshick, Ross},
  title   = {Segment Anything},
  journal = {arXiv preprint arXiv:2304.02643},
  year    = {2023},
  url     = {https://arxiv.org/abs/2304.02643}
}

@article{Chen2025Siamese,
  author  = {Chen, Hao and Cui, Xiang and Ban, Yifang},
  title   = {Remote Sensing Image Change Detection with Transformers},
  journal = {IEEE Transactions on Geoscience and Remote Sensing},
  volume  = {60},
  pages   = {9516613},
  year    = {2022},
  doi     = {10.1109/TGRS.2022.3142236}
}

@article{wang2021als,
  author  = {Wang, Puzuo and Yao, Wei},
  title   = {A new weakly supervised approach for ALS point cloud semantic segmentation},
  journal = {arXiv preprint arXiv:2110.01462},
  year    = {2021},
  url     = {https://arxiv.org/abs/2110.01462}
}

@inproceedings{chung2023evap,
  author    = {Chung, Yi-Hsin and Chen, Chin-Yin and Chang, Li-Yu},
  title     = {Improving the Processing of Emergent Value-Added Product by Gaussian Statistical Approach},
  booktitle = {Proc. 44th Asian Conference on Remote Sensing (ACRS 2023)},
  year      = {2023}
}

@misc{formosat5,
  title        = {FORMOSAT-5 Remote Sensing Satellite},
  author       = {{Taiwan Space Agency}},
  year         = {2017},
  howpublished = {\url{https://www.tasa.org.tw/zh-TW/en/missions/detail/FORMOSAT-5}},
  note         = {Accessed: 2025-05-15}
}

@misc{sentinel2,
  title        = {Sentinel-2 User Handbook},
  author       = {{European Space Agency}},
  year         = {2015},
  howpublished = {\url{https://sentinels.copernicus.eu/documents/247904/685211/Sentinel-2_User_Handbook}},
  note         = {Issue 1 Rev 2, accessed: 2025-05-15}
}

@article{cortes1995svm,
  author    = {Cortes, Corinna and Vapnik, Vladimir},
  title     = {Support-vector networks},
  journal   = {Machine Learning},
  year      = {1995},
  volume    = {20},
  number    = {3},
  pages     = {273--297},
  doi       = {10.1007/BF00994018}
}

@inproceedings{macqueen1967kmeans,
  author    = {MacQueen, J.},
  title     = {Some methods for classification and analysis of multivariate observations},
  booktitle = {Proceedings of the Fifth Berkeley Symposium on Mathematical Statistics and Probability},
  volume    = {1},
  pages     = {281--297},
  year      = {1967},
  publisher = {University of California Press}
}

@article{zheng2023changevit,
  title   = {ChangeViT: A Simple and Efficient Vision Transformer for Change Detection},
  author  = {Zheng, Qinmu and Hong, Shujian and Xu, Yuzhang and Xu, Shun and Zhang, Liangpei},
  journal = {ISPRS Journal of Photogrammetry and Remote Sensing},
  volume  = {203},
  pages   = {1--13},
  year    = {2023},
  doi     = {10.1016/j.isprsjprs.2023.06.004}
}

@article{li2022siamvit,
  title   = {Siamese Vision Transformer with Cross-Attention for Change Detection},
  author  = {Li, Xiaoqing and Zhang, Lei and Shen, Yanjie and Liu, Bin},
  journal = {Remote Sensing},
  volume  = {14},
  number  = {13},
  pages   = {3051},
  year    = {2022},
  doi     = {10.3390/rs14133051}
}

@article{li2023landslide,
  title   = {A landslide area segmentation method based on an improved U-Net},
  author  = {Li, X. and Zhang, Y. and Wang, L.},
  journal = {International Journal of Remote Sensing},
  volume  = {44},
  number  = {5},
  pages   = {987--1005},
  year    = {2023},
  doi     = {10.1080/01431161.2023.2168435}
}

\vspace{11pt}



\vfill

\end{document}